\documentclass[10pt,twocolumn,letterpaper]{article}

\usepackage[pagenumbers]{cvpr} % To force page numbers, e.g. for an arXiv version
\usepackage{fontawesome5}
\usepackage{gradient-text}
\usepackage{subcaption}

\definecolor{cvprblue}{rgb}{0.21,0.49,0.74}
\usepackage[pagebackref,breaklinks,colorlinks,allcolors=cvprblue]{hyperref}
\usepackage[table]{xcolor}
\usepackage[most]{tcolorbox}
\usepackage{xcolor}
\usepackage[table]{xcolor} 
\definecolor{hos}{HTML}{E97132} 
\definecolor{hps}{HTML}{A02B93}
\newtcolorbox{hypobox}{
  colback=blue!5!white,  
  colframe=blue!40!black, 
  boxrule=0.6pt,          
  arc=2mm,                
  left=6pt, right=6pt, top=4pt, bottom=4pt,
  before skip=10pt, after skip=10pt
}
\newtcolorbox{conbox}{
  colback=green!5!white,   
  colframe=green!40!black, 
  boxrule=0.6pt,
  arc=2mm,
  left=6pt, right=6pt, top=4pt, bottom=4pt,
  before skip=10pt, after skip=10pt
}
\newtcolorbox{quesbox}{
  colback=orange!8!white,
  colframe=orange!60!black,
  boxrule=0.6pt,
  arc=2mm,
  left=6pt, right=6pt, top=4pt, bottom=4pt,
  before skip=10pt, after skip=10pt
}
\newtcolorbox{redbox}{
  colback=gray!10!white,
  colframe=red!60!black,
  boxrule=0.6pt,
  arc=2mm,
  left=6pt, right=6pt, top=3pt, bottom=3pt,
  before skip=10pt, after skip=10pt
}
\newtcolorbox{greenbox}{
  colback=gray!10!white,
  colframe=green!60!black,
  boxrule=0.6pt,
  arc=2mm,
  left=6pt, right=6pt, top=3pt, bottom=3pt,
  before skip=10pt, after skip=10pt
}
\newtcolorbox{bluebox}{
  colback=gray!10!white,
  colframe=blue!60!black,
  boxrule=0.6pt,
  arc=2mm,
  left=6pt, right=6pt, top=3pt, bottom=3pt,
  before skip=10pt, after skip=10pt
}

\definecolor{tableblue}{RGB}{201,226,239}
\definecolor{tableorange}{RGB}{255,239,213}
\definecolor{tablepink}{RGB}{255,240,245}
\definecolor{tablegray}{RGB}{240,240,240}
\definecolor{tableyellow}{RGB}{255,253,231} 
\definecolor{tablegreen}{RGB}{232,245,233}
\definecolor{tablemint}{RGB}{224,255,255}

\definecolor{hosbg}{HTML}{BBD4EE}   
\definecolor{hpsbg}{HTML}{B9E0DE}   
\definecolor{headerline}{HTML}{6D8FB4}

\definecolor{TakeawayRed}{RGB}{200, 82, 82}      
\definecolor{TakeawayBlue}{RGB}{30, 80, 240}    
\definecolor{TakeawayGreen}{RGB}{34, 139, 34}   
\definecolor{TakeawayPurple}{RGB}{155, 89, 182}  
\definecolor{TakeawayOrange}{RGB}{230, 126, 34}  
\definecolor{TakeawayTeal}{RGB}{22, 160, 133}    

\newcommand{\hvs}{\textbf{\textit{HVS}}}
\newcommand{\hos}{\textbf{\textit{HOS}}}
\newcommand{\hps}{\textbf{\textit{HPS}}}
\newcommand{\hstar}{\textbf{\textit{H$^*$Bench}}}

\title{\textit{Thinking in 360°}: Humanoid Visual Search in the Wild}

\author{
  Heyang Yu\textsuperscript{1}\thanks{Equal contribution.} \quad
  Yinan Han\textsuperscript{3}\footnotemark[1] \quad
  Xiangyu Zhang\textsuperscript{4} \quad
  Baiqiao Yin\textsuperscript{1} \\
  Bowen Chang\textsuperscript{1} \quad
  Xiangyu Han\textsuperscript{1} \quad
  Xinhao Liu\textsuperscript{1} \quad
  Jing Zhang\textsuperscript{1} \\
  Marco Pavone\textsuperscript{2,5} \quad
  Chen Feng\textsuperscript{1}\footnotemark[2] \quad
  Saining Xie\textsuperscript{1}\footnotemark[2] \quad
  Yiming Li\textsuperscript{1,2}\footnotemark[2] \vspace{2mm} \\ 
  $^{1}$NYU \quad 
  $^{2}$NVIDIA \quad 
  $^{3}$TU Darmstadt \quad 
  $^{4}$UC Berkeley \quad 
  $^{5}$Stanford University \\
  {\small \url{https://humanoid-vstar.github.io}}
  \vspace{-3mm}
}

\begin{document}
\twocolumn[{
    \renewcommand\twocolumn[1][]{#1}
    \maketitle
    \vspace{-3em}
    \begin{center}
        \centering
        \includegraphics[width=0.86\textwidth]{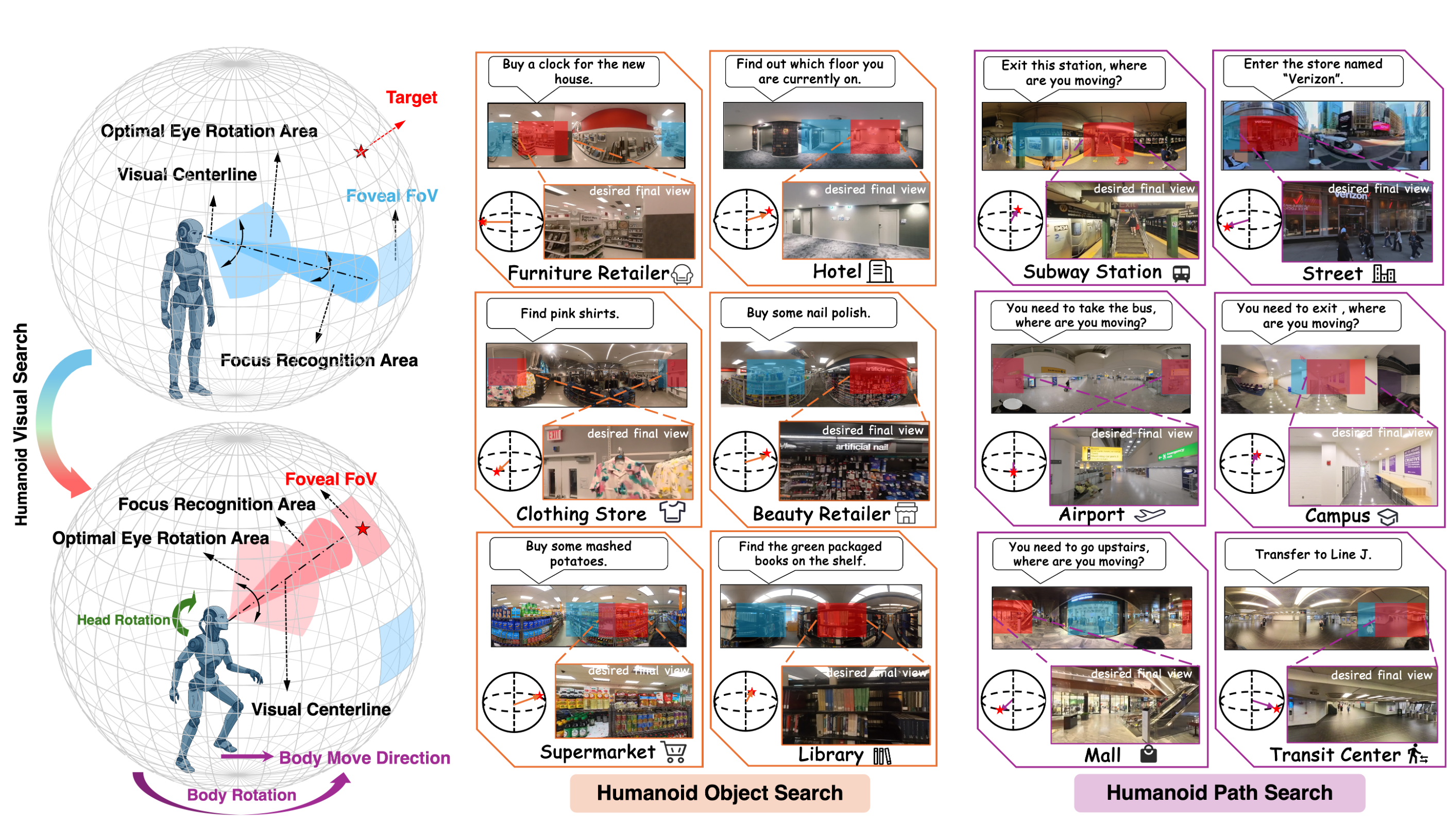}
        \vspace{-3mm}
        \captionof{figure}{\textbf{We pose a fundamental question: \textit{can an AI agent actively search for objects or paths in a 3D world like a human, rather than just passively describe what it sees?}} We transition agents from passive captioners to active searchers, moving them from constrained household scenes into the rich complexity of the wild. A key enabler is our scalable paradigm where a single 360° panorama closes the perception-action loop for head rotation, freeing active embodied reasoning in real life from hardware constraints.}
        \vspace{-1mm}
        \label{fig:teaser}
    \end{center}
}]

\footnotetext[1]{Equal contribution.} 
\footnotetext[2]{Equal advising.}

\begin{abstract}
% \vspace{-1mm}
Humans rely on the synergistic control of head (cephalomotor) and eye (oculomotor) to efficiently search for visual information in 360°. However, prior approaches to visual search are limited to a static image, neglecting the physical embodiment and its interaction with the 3D world. How can we develop embodied visual search agents as efficient as humans while bypassing the constraints imposed by real-world hardware? 
To this end, we propose \textbf{humanoid visual search} where a humanoid agent actively rotates its head to search for \textbf{objects} or \textbf{paths} in an immersive world represented by a 360° panoramic image. To study visual search in visually-crowded real-world scenarios, we build \hstar, a new benchmark that moves beyond household scenes to challenging in-the-wild scenes that necessitate advanced visual-spatial reasoning capabilities, such as transportation hubs, large-scale retail spaces, urban streets, and public institutions. Our experiments first reveal that even top-tier proprietary models falter, achieving only \textbf{$\sim$30\%} success in object and path search. We then use post-training techniques to enhance the open-source Qwen2.5-VL, increasing its success rate by over \textbf{threefold} for both object search (14.83\% → 47.38\%) and path search (6.44\% → 24.94\%). Notably, the lower ceiling of path search reveals its inherent difficulty, which we attribute to the demand for sophisticated spatial commonsense. Our results not only show a promising path forward but also quantify the immense challenge that remains in building MLLM agents that can be seamlessly integrated into everyday human life. 
\vspace{-5mm}
\end{abstract}
    
\section{Introduction}
\label{intro}
The human visual system is highly efficient, capturing sharp detail only at the fovea while leaving other regions blurred or unseen~\cite{marr2010vision}. Despite the sensor limitations, humans efficiently perform visual search tasks in 360° (\eg, locating the next exit in a crowded subway station) by rapidly executing saccades and deliberately reorienting their head, minimizing sensory redundancy and computational cost.

Recent state-of-the-art computational methods for visual search are based on Multimodal Large Language Models (MLLMs), leveraging their rich world knowledge to enhance the generalization capability. These methods commonly process a single, static, low-resolution image. Subsequent actions are typically confined to computational operations (\eg, cropping and zooming within this fixed canvas) to enhance resolution and glean detail~\citep{wu2024v,man2025argus,zhangmllms}. However, these methods suffer from two fundamental gaps compared to \textit{biological visual search}~\citep{bartz1966eye}: \textbf{(1)~Non-interactive}: Lacking an interactive simulator, the model cannot change its perspective to obtain information beyond its initial field of view. \textbf{(2)~Disembodied}: Lacking physical embodiment, the model cannot couple visual reasoning with actions in the physical world. Moreover, the search is usually not driven by embodied tasks (\eg, manipulation or navigation), reducing it to an abstract perceptual exercise rather than a real-world, goal-directed behavior~\cite{li2025dyfo, zhang2025chain, su2025pixel}. In other words, \textit{developing an embodied visual agent that can actively search for information in visually crowded scenes remains underexplored, despite its significant potential in humanoid robots, assistive technology, and augmented reality.}

To bridge these gaps, we prototype \textit{\textbf{humanoid visual search (HVS)}}, where humanoid agents couple deliberate reasoning with active head turns for visual search in complex environments. Specifically, \textbf{(1)~\hvs~ is interactive}. The agent starts with a narrow perspective view but acts within a lightweight 360° panorama, enabling a closed-loop perception–action cycle in which each head rotation changes its visual input. \textbf{(2)~\hvs~ is embodied}. It tightly couples visual reasoning with physical action, requiring agents to deliberately coordinate head movements as part of their thought process. 
Meanwhile, the search is driven by embodied tasks, defined in two core forms: \textbf{\textit{a.~humanoid object search (HOS)}}: Locating and foveating a target object as a prerequisite for manipulation. \textbf{\textit{b.~humanoid path search (HPS)}}: Identifying a navigable path to a destination and aligning body orientation as a prerequisite for locomotion. To systematically study \hvs, we raise the question: 
\begin{quote}
   \textit{ What environments truly necessitate advanced visual-spatial reasoning capabilities?}
\end{quote}
 We argue that human-made environments rich in \textcolor{TakeawayRed}{\textbf{\textit{structural}}} (multi-level layouts), \textcolor{TakeawayGreen}{\textbf{\textit{semantic}}} (dense compositional cues), and \textcolor{TakeawayBlue}{\textbf{\textit{volumetric}}} (cluttered 3D space) complexities offer the most valuable testbed for such reasoning. We therefore \textbf{\textit{shift our focus beyond simplistic constrained scenarios such as household object search toward in-the-wild challenges}}, such as: \textit{navigating the multi-level labyrinth of a subway station to find a specific exit, identifying a particular store in a large shopping mall, or retrieving a specific product from a densely stocked supermarket aisle.} Unfortunately, existing embodied AI platforms often suffer from limited perceptual realism~\citep{dosovitskiy2017carla,bruce2024genie,wu2025metaurban} or are restricted to household scenes~\citep{xia2018gibson,savva2019habitat,deitke2022,li2023behavior}, failing to represent the rich, dense, and cluttered environments that necessitate advanced visual-spatial reasoning capabilities. 

To this end, we introduce~\hstar, a new benchmark featuring diverse in-the-wild scenes suitable for \hvs, including but not limited to transportation hubs (airports and subway stations), large-scale retail spaces (supermarkets and shopping malls), and public institutions (libraries and museums), as shown in Fig.~\ref{fig:teaser}. Each panoramic image is densely annotated with embodied task questions and corresponding ground-truth actions: optimal head orientations for \hos~and unit direction vectors on the ground plane for \hps. We evaluate MLLMs in their ability to focus on task-driven visual details and generate embodied plans in diverse human environments. Our experiments show that supervised fine-tuning (SFT) and reinforcement learning (RL) can enhance performance, yet also reveal major unresolved challenges, highlighting the long-term value of research in this domain. Our contributions are listed as follows.
\begin{itemize}
    \item We introduce humanoid visual search, a \textbf{novel task} enabling human-like active spatial reasoning in 360°.
    \item We propose a \textbf{scalable framework} that leverages real-world 360° panoramas as {lightweight simulators}, creating a hardware-free platform to study embodied reasoning.
    \item We build \textit{H$^*$Bench}, the first \textbf{systematic benchmark} of its kind, with dense annotations on panoramic scenes from challenging in-the-wild environments.
    \item We conduct \textbf{thorough evaluations} to show that post-training can improve the performance of MLLMs, while also highlighting promising avenues for future work.
\end{itemize}

\section{Related Works}
\label{related}

\noindent \textbf{Visual Search.} As a hallmark of \textit{System 2} slow thinking~\citep{kahneman2011thinking}, visual search involves deliberate reasoning to identify objects or information within visually crowded environments.  Early visual search methods use bottom-up visual saliency, top-down contextual guidance, or a combination of both~\citep{oliva2003top,zelinsky2005role,torralba2006contextual}. Yet these studies often fail to generalize due to their limitations in contextual understanding. Recent advances in visual search, pioneered by~\textit{V$^*$}~\citep{wu2024v}, are driven by MLLMs with rich world knowledge (\textit{e.g.}, object co-occurrence) to achieve better performance. However, \textit{existing work focuses on search within static 2D images, without considering the active and embodied nature of visual search in the 3D world}. Prior neuroscience studies reveal that human visual search coordinates eyes and head in a nested system, in which the head preferentially explores unseen regions, while the eyes exploit already-visible content via finer-scale saccades~\citep{lanman1978coordination,barnes1979vestibulo,solman2017eye}. \textit{We prototype a visual search model with human-like eye-head coordination.}

% We make the first effort to reverse engineer this fundamental cognitive hierarchy.
% MLLMs represent a breakthrough in this regard, as their capacity for chained reasoning allows them to formulate explicit exploratory goals and then ground them in detailed, exploitative perceptual acts, providing the first viable model of this fundamental cognitive hierarchy. 

\noindent\textbf{Visual Navigation.} Visual navigation, along with vision-language navigation, aims to develop agents that can move through an environment to reach a specified goal~\cite{zhu2017target,anderson2018vision}. The core of these tasks lies in completing the entire trajectory as fast as possible. However, this requires a 3D simulator or real hardware: realistic simulators are hard to build, and real-world experiments are difficult to scale and reproduce. As a result, previous efforts are mostly limited to household scenes where 3D data is easier to acquire~\cite{chaplot2020neural,chaplot2020learning,chaplot2020object,chang2024goat,majumdar2022zson,zhou2024navgpt,qi2025vln}, leaving in-the-wild challenges underexplored. Our work is motivated by the observation that \textit{human reasoning is intermittent during navigation}; it is invoked only at critical decision points. Focusing on these critical points enables us to build a closed-loop search environment directly from in-the-wild 360° panoramas, bypassing the need for 3D simulation or physical hardware and yielding a scalable framework for embodied visual search. In summary, \textit{we focus on active, embodied, and multimodal reasoning in visually crowded 360° scenes, a key prerequisite for open-world navigation and mobile manipulation.}

\noindent\textbf{Multimodal LLMs.}
Multimodal LLMs can understand and reason about multiple modalities of information (\textit{e.g.}, text and images), representing a promising pathway toward Artificial General Intelligence (AGI). Seminar works focus on effectively aligning the feature spaces of pre-trained visual encoders with those of LLMs, including Flamingo~\citep{alayrac2022flamingo}, BLIP~\citep{li2022blip,li2023blip}, and LLaVA~\citep{liu2023llava}.  Recently, \citet{tong2024cambrian} conduct a systematic vision-centric exploration and propose a token-efficient connector to integrate high-resolution vision
features with LLMs. Meanwhile, a new wave of MLLMs, such as GPT-4o~\citep{gpt4o_blog} and Gemini~2.5~\citep{sadigh2025gemini25pushing}, has set new state-of-the-art benchmarks across multimodal tasks by scaling model capacity and introducing novel training recipes. One crucial technique is reinforcement learning (RL)-based post-training, which further aligns model outputs with human preferences, enhances instruction-following capabilities, and mitigates hallucination and bias~\cite{ouyang2022training,yu2024rlhf}. RL can also incentivize strong reasoning abilities when coupled with verifiable rewards, particularly for complex tasks such as mathematical problem solving and multistep logical reasoning~\citep{guo2025deepseek,huang2025vision,team2025kimik15,team2025kimivl}. \textit{We ground MLLMs in the physical world to assess and improve their performance in active and embodied visual search.}

\noindent\textbf{Multimodal LLMs with Tools.}
Just as humans leverage external tools to transcend their innate physical and cognitive limitations, a similar paradigm is now empowering artificial intelligence. Recently, LLM agents demonstrate superior performance in solving challenging long-horizon tasks~\citep{jin2025search, OpenManus}, a capability unlocked by providing them with an external toolkit (\eg, web browsing, code execution) and refining their policies through multi-turn reinforcement learning~\citep{fu2025areal,feng2025group}. This approach is now being extended to multimodal settings, where MLLMs generate a symbolic tool call  (\textit{e.g.}, OCR, marking, cropping, zoom in) at each reasoning step to overcome its limitations in semantic grounding and visual perception~\citep{zhang2025chain,qi2025cogcom,zheng2025deepeyes,shen2024zoomeye,liu2024chain}. Among these tools, iterative zoom-in and region-of-interest selection are particularly well-suited for visual search tasks, enabling active perception over images. However, these operations still occur on a disembodied 2D canvas, where the tool (action) is confined to computational manipulations of a static image file. \textit{We couple the use of the tool with the actions in the physical world: active head rotation is called to continuously construct a visual chain of thought. This bridges the critical gap between passive visual reasoning and active embodied reasoning.}

\noindent\textbf{Multimodal LLMs for Embodied Reasoning.} While LLMs trained on Internet data may acquire rich world knowledge, they struggle to ground that knowledge in the physical world due to the large gap between symbolic linguistic representations and embodied perception. To bridge the gap, a growing body of research aims to ground MLLMs in embodied reasoning~\citep{tian2025drivevlm,hwang2024emma,zawalski2025robotic,zhao2025cot, Chen25-ecot-lite}. To build more general models, Cosmos-Reason1~\citep{azzolini2025cosmos} enables MLLMs to perceive the physical world through video input and reason about it to generate more physically grounded responses, including explanatory insights and embodied decisions.  Gemini Robotics-ER~\citep{team2025gemini} extends Gemini’s multimodal reasoning capabilities to the physical world, with
enhanced spatiotemporal understanding. \textit{Yet active visual search with interleaved multimodal reasoning remains unexplored.}

\section{Humanoid Visual Search}
\label{method}

Human spatial intelligence is punctuated by critical decision points where we stop to observe, reason, and resolve ambiguity before acting with confidence. Our task focuses on these critical points, abstracting full-body motion into the \textit{atomic action} of head rotation. This is motivated by the crucial role of cephalomotor control in human visual search, allowing us to study the core cognitive processes of embodied visual search in a tractable yet realistic manner. 

\begin{figure*}[t]
    \begin{center}
        \includegraphics[width=0.9\textwidth]{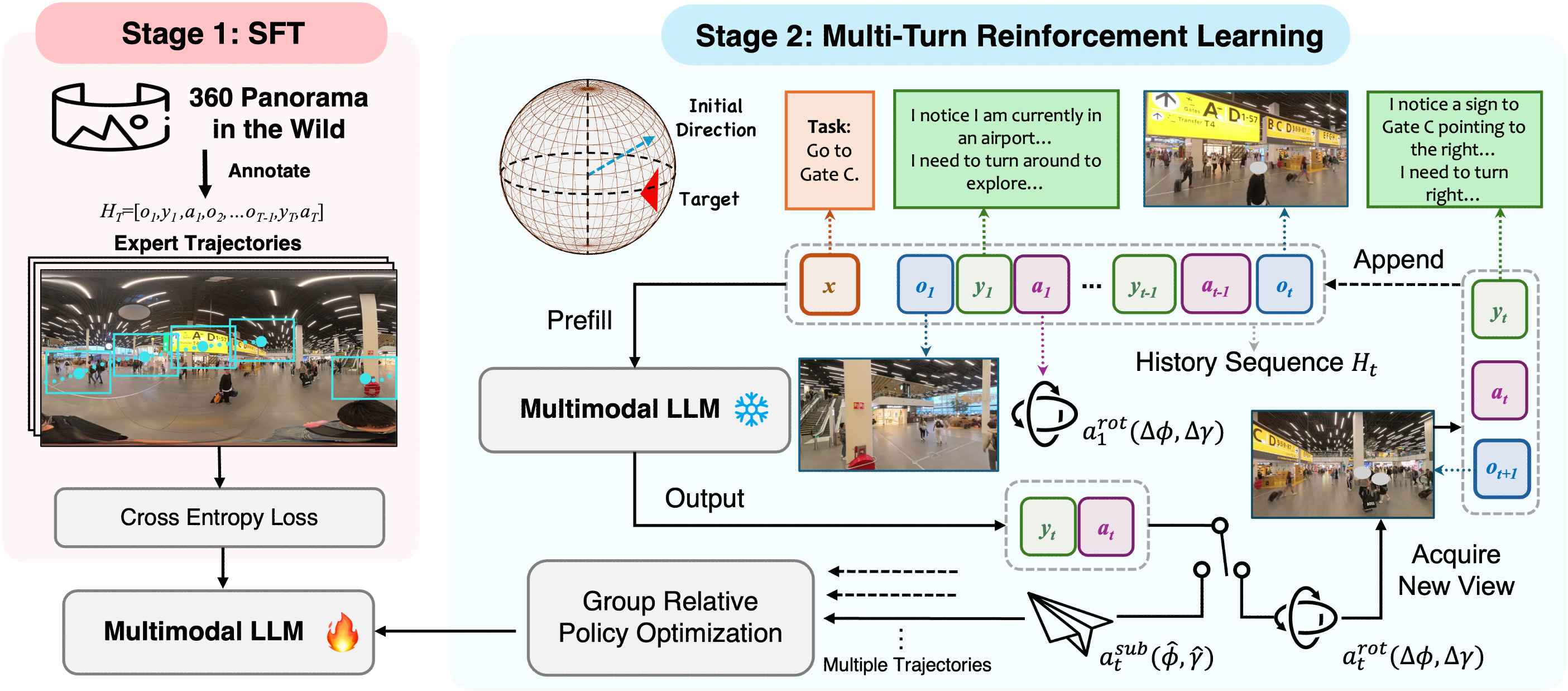} 
        \vspace{-1mm}
        \caption{\textbf{Pipeline Illustration.} Stage 1 (SFT) provides the foundational ability to map perspective images to plausible actions (\eg, turning around upon seeing nothing). Stage 2 (RL) refines this into a strategic policy: the model learns to explore (outputting $a_t^{rot}$) until it acquires a view with sufficient evidence (\eg, spotting the gate sign), at which point it confidently submits the final estimate ($a_t^{sub}$).} 
        \vspace{-6mm}
        \label{fig:2stage}
    \end{center}
\end{figure*}
% \vspace{-2mm}
\subsection{Problem Formulation}
\label{hvs}
\paragraph{Objective.} Imagine a humanoid agent equipped with a limited field-of-view (FoV) navigating a complex, multi-corridor junction in a subway station, tasked with locating an exit as quickly as possible. The limited FoV necessitates a tight coordination between head and eye movements:  the head explores the unknown by rotating to new vantage points, while the eyes exploit the already-seen information by fixating on task-relevant details. Formally, we model the environment as a single 360$^\circ$ panoramic image. The set of all possible observations, $\mathcal{S}_o = \{ o_{\phi,\gamma} \}$, comprises narrow-FoV perspective images sampled from this panorama, each defined by its azimuth ($\phi$) and polar angle ($\gamma$). The goal of \hvs\ is to identify the optimal direction $(\phi^*,\gamma^* )$ that maximizes the probability of task success $r_s$ given the language instruction $x$ and visual observation $o_{\phi,\gamma}$: 
$$
(\phi^*,\gamma^*) = \arg \max_{\phi,\gamma} P(r_s \mid o_{\phi,\gamma},x).
$$

\noindent\textbf{Humanoid Object Search (HOS).} HOS tackles the problem of active target search in an unknown 3D environment by finding a final viewing direction $(\phi^*,\gamma^*)$ that brings the target into the central foveal region of the perspective view.

\noindent\textbf{Humanoid Path Search (HPS).} HPS requires the agent to search for a navigable path to a target location as a high-level planning step before locomotion. The goal is to identify a final viewing direction $\phi^*$ that is aligned with the path.

\subsection{Humanoid Visual Search with MLLMs}
We frame humanoid visual search as a multimodal reasoning task by coupling MLLM tool use with head rotation. This is realized through a tool-augmented MLLM~\citep{he2025thinkingimagesmultimodal}, with the agent policy defined as $\pi_\theta (y_t, a_t\mid o_t, x, \mathcal{H}_t)$. Specifically, at each timestep $t$, the agent generates a textual chain of thought $y_t$ and an action $a_t$, conditioned on current observation $o_t = o_{\phi_t,\gamma_t}$, language instruction $x$ and history states $\mathcal{H}_t = \{(o_i, y_i, a_i)\}_{i=1}^{t-1}$. Each search episode allows for a sequence of rotation actions, concluding with a submission action that yields the final output. The action space thus consists of two primitives:
\begin{itemize}
    \item \textbf{Rotate} $a_t^{rot} = (\Delta \phi, \Delta \gamma)$: Adjusts the viewing direction, updating $\phi_{t+1} = \phi_t + \Delta \phi$ and $\gamma_{t+1} = \gamma_t + \Delta \gamma$ (right/up are positive; yaw is circular).
    \item \textbf{Submit} $a_t^{sub}$: Commits the current viewing direction as the final estimate $(\hat{\phi},\hat{\gamma})$ and terminates the episode.
\end{itemize}

\subsection{MLLM Post-Training}
Trained on static, disembodied Internet data, MLLMs inherently lack the spatial commonsense and active 3D planning capabilities required for humanoid visual search. Our evaluations (\cref{subsec:exp:probing}) reveal that even state-of-the-art proprietary models like GPT-4o~\citep{gpt4o_blog} achieve only $\sim$20\% success. Hence, we adapt MLLMs into effective visual search agents through the two-stage post-training pipeline shown in \cref{fig:2stage}. We outline the stages below while deferring detailed mathematical formulations to Appendix~\cref{objective func}.

\noindent\textbf{Stage 1: Supervised Fine-Tuning (SFT).}
We first perform SFT on a curated multi-turn dataset (\cref{subsec:bench:bench}) to instill basic task-orientied reasoning and tool-use abilities. This teaches the model to generate structured action plans from multimodal inputs, establishing a strong behavioral prior.

\noindent\textbf{Stage 2: Multi-Turn Reinforcement Learning (RL).}
We then refine the policy using Group Relative Policy Optimization (GRPO)~\citep{Shao2024DeepSeekMathPT}. This RL stage encourages long-horizon reasoning and is crucial for developing robust, generalizable search strategies beyond the imitation learning baseline according to the previous findings~\cite{chusft}.

\begin{figure*}[t]
    \begin{center}
        \includegraphics[width=\textwidth]{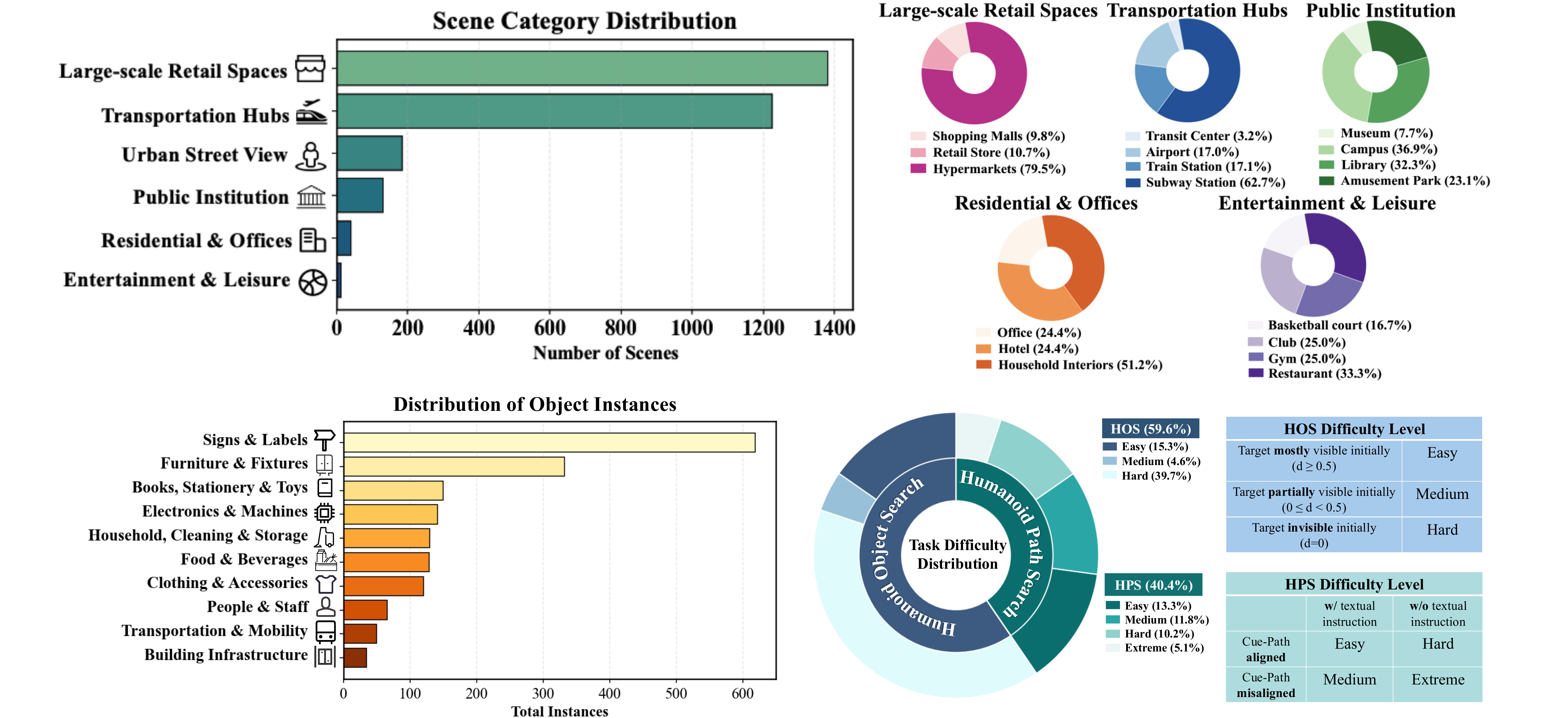}
        % \vspace{-2mm}
        \caption{\textbf{Top Left:} Distribution of scene categories. \textbf{Top Right:} Sub-category composition per category. \textbf{Bottom Left:} Category-wise object instance distribution for \textbf{\textit{HOS}}. \textbf{Bottom Right:} Task difficulty distribution and definition for \textbf{\textit{HOS}} and \textbf{\textit{HPS}}.}
        \vspace{-5mm}
        \label{fig:combined_chart.png}
    \end{center}
\end{figure*}
\vspace{-2mm}
\section{H$^*$Bench}\label{sec:bench}
\subsection{Dataset Overview}\label{subsec:bench:data}
We introduce \hstar~to systematically evaluate humanoid visual search in rich, dense, and visually cluttered real-world environments. The benchmark comprises approximately \textbf{3,000} annotated task instances derived from a diverse set of high-resolution panoramic videos (up to $7680 \times 3840$). By initializing the agent with four distinct starting orientations per task instance, we obtain a total of \textbf{12,000} search episodes. Sourced from both self-collected footage across global metropolitan areas (New York, Paris, Amsterdam, Frankfurt) and open platforms (YouTube and the 360+x dataset~\citep{chen2024x360}), \hstar~delivers broad geographical coverage (see Appendix \cref{geographical_distribution}) and substantial scene diversity (\cref{fig:combined_chart.png} Top). More specifically, it spans a wide spectrum of challenging scenarios across \textbf{12} countries, systematically organized into \textbf{6} major scene categories and \textbf{18} fine-grained scene types. These range from densely stocked retail environments and bustling transportation hubs to diverse public institutions and urban streets. Additionally, the diversity of \hos~target objects is summarized in Fig.~\ref{fig:combined_chart.png} Bottom Left. This breath across scenes and targets provides a rigorous and comprehensive evaluation testbed for assessing the visual search abilities of humanoid agents operating in complex real-world environments.

\begin{table*}[t]
  \centering
  \caption{\textbf{ Left:} Quantitative results of open-source, proprietary, and fine-tuned models on \hstar. Top-three performances are highlighted with {\colorbox{red!7}{red}}, {\colorbox{Green!7}{green}} and {\colorbox{blue!7}{blue}}. \textbf{Right:} Performance comparison of the best-in-class open-source, proprietary, and fine-tuned models.} 
  \label{tab:bench}
  % \vspace{-2mm}
  \begin{minipage}[t]{0.7\textwidth}
    \vspace{0pt}
    \centering
    \resizebox{\linewidth}{!}{%
      \begin{tabular}{lcccccccccc}
      \toprule
           % & \multicolumn{4}{c}{\cellcolor{hos!20}\textbf{\hos}} & & \multicolumn{5}{c}{\cellcolor{hps!20}\textbf{\hps}} 
           & \multicolumn{4}{c}{\textbf{Humanoid Object Search}} & & \multicolumn{5}{c}{\textbf{Humanoid Path Search}}
           \\
          % \cline{2-5}\cline{7-11}
             \textbf{Method} & \textbf{Overall} & \textbf{Easy} & \textbf{Medium} & \textbf{Hard} & & 
             \textbf{Overall} & \textbf{Easy} & \textbf{Medium} & \textbf{Hard} & \textbf{Extreme} \\
          \hline
          \rowcolor{tablegray}
          \hspace{-0.3em}\textit{Open-Weight Multi Image Models} & & & & & & & & & & \\
          InternVL3.5-4B~\citep{chen2025internvl35advancingopensource} & 3.08 & 7.32 & 2.84 & 1.49 & & 4.81 & 6.00 & 5.70 & 4.67 & 0.46 \\
          InternVL3.5-8B~\citep{chen2025internvl35advancingopensource} & 6.38 & 9.76 & 9.10 & 4.79 & & 7.25 & 10.00 & 7.68 & 5.14 & 4.17 \\
          Qwen2.5-VL-3B-Instruct~\citep{xu2025qwen25vltechnicalreport} & 14.83 & 27.97 & 13.07 & 10.01 & & 6.44 & 7.00 & 8.77 & 4.91 & 3.24 \\
          Qwen2.5-VL-7B-Instruct~\citep{xu2025qwen25vltechnicalreport} & 11.38 & 23.42 & 9.10 & 7.02 & & 6.31 & 9.00 & 5.92 & 5.84 & 1.85 \\
          Gemma-3-4B-it~\citep{gemmateam2025gemma3technicalreport} &  17.13 & 32.85 &\cellcolor{red!7}\textbf{26.14} & 10.13 & & 14.44 & 17.20 & 14.47 & 14.72 & 7.41 \\
          Gemma-3-12B-it~\citep{gemmateam2025gemma3technicalreport} & 10.21 & 24.72 & 17.33 & 3.88 & & 14.50 & 16.80 & 14.25 & 14.49 &9.72 \\
          Kimi-VL-A3B-Instruct~\citep{kimiteam2025kimivltechnicalreport} & 4.92 & 12.85 & 0.57 & 2.36 & & 4.32 & 8.79 & 3.32 & 2.21 & 4.17 \\
          \hline
          \rowcolor{tablegray}
          \hspace{-0.3em}\textit{Proprietary Models} & & & & & & & & & & \\
          GPT-4o~\citep{gpt4o_blog} & 19.75 & 18.17 & 17.35 & 20.92 & & \cellcolor{blue!7}\textbf{23.69} & 26.80 & \cellcolor{blue!7}\textbf{22.59} & \cellcolor{green!7}\textbf{26.17} & \cellcolor{blue!7}\textbf{13.89} \\
          Gemini2.5-Pro~\citep{sadigh2025gemini25pushing} & \cellcolor{blue!7}\textbf{31.96} &\cellcolor{blue!7}\textbf{33.58} & 23.78 & \cellcolor{blue!7}\textbf{32.13} & &\cellcolor{red!7}\textbf{33.00} & \cellcolor{red!7}\textbf{41.60} & \cellcolor{red!7}\textbf{29.39} & \cellcolor{red!7}\textbf{35.75} & \cellcolor{red!7}\textbf{15.28} \\
          \hline
          \rowcolor{tablegray}
          \hspace{-0.3em}\textit{Fine-Tuned Models \textbf{(Ours)}} & & & & & & & & & & \\
          % \textbf{HVS-3B-mixed-rl-w/osft} & 6.58 & 9.25 & 4.55 & 5.79 & & 0.00 & 0.00 & 0.00 & 0.00 & 0.00 \\
          \textbf{HVS-3B (w/ SFT only)}\ & \cellcolor{green!7}\textbf{40.83} & \cellcolor{green!7}\textbf{53.82} & \cellcolor{blue!7}\textbf{23.86} & \cellcolor{green!7}\textbf{37.73} & & 23.00 & \cellcolor{blue!7}\textbf{28.00} & \cellcolor{green!7}\textbf{23.03} & 21.26 & \cellcolor{green!7}\textbf{14.81} \\
          % \textbf{HVS-3B-mixed-sft-w/ rl}\  &\cellcolor{red!5}\textbf{47.38} & \cellcolor{red!7}\textbf{60.49} & \cellcolor{green!7}\textbf{24.43} & \cellcolor{red!7}\textbf{44.87} & & \cellcolor{green!7}\textbf{24.94} & \cellcolor{green!7}\textbf{34.80} & 20.18 & \cellcolor{blue!7}\textbf{25.00} & 12.04 \\
          \textbf{HVS-3B}\  &\cellcolor{red!5}\textbf{47.38} & \cellcolor{red!7}\textbf{60.49} & \cellcolor{green!7}\textbf{24.43} & \cellcolor{red!7}\textbf{44.87} & & \cellcolor{green!7}\textbf{24.94} & \cellcolor{green!7}\textbf{34.80} & 20.18 & \cellcolor{blue!7}\textbf{25.00} & 12.04 \\
          \bottomrule
      \end{tabular}
    }% end resizebox
  \end{minipage}
  \hfill
  % ===== 右侧：图 0.23 =====
  \begin{minipage}[t]{0.27\textwidth}
    \vspace{0pt}
    \centering
    \includegraphics[width=\linewidth]{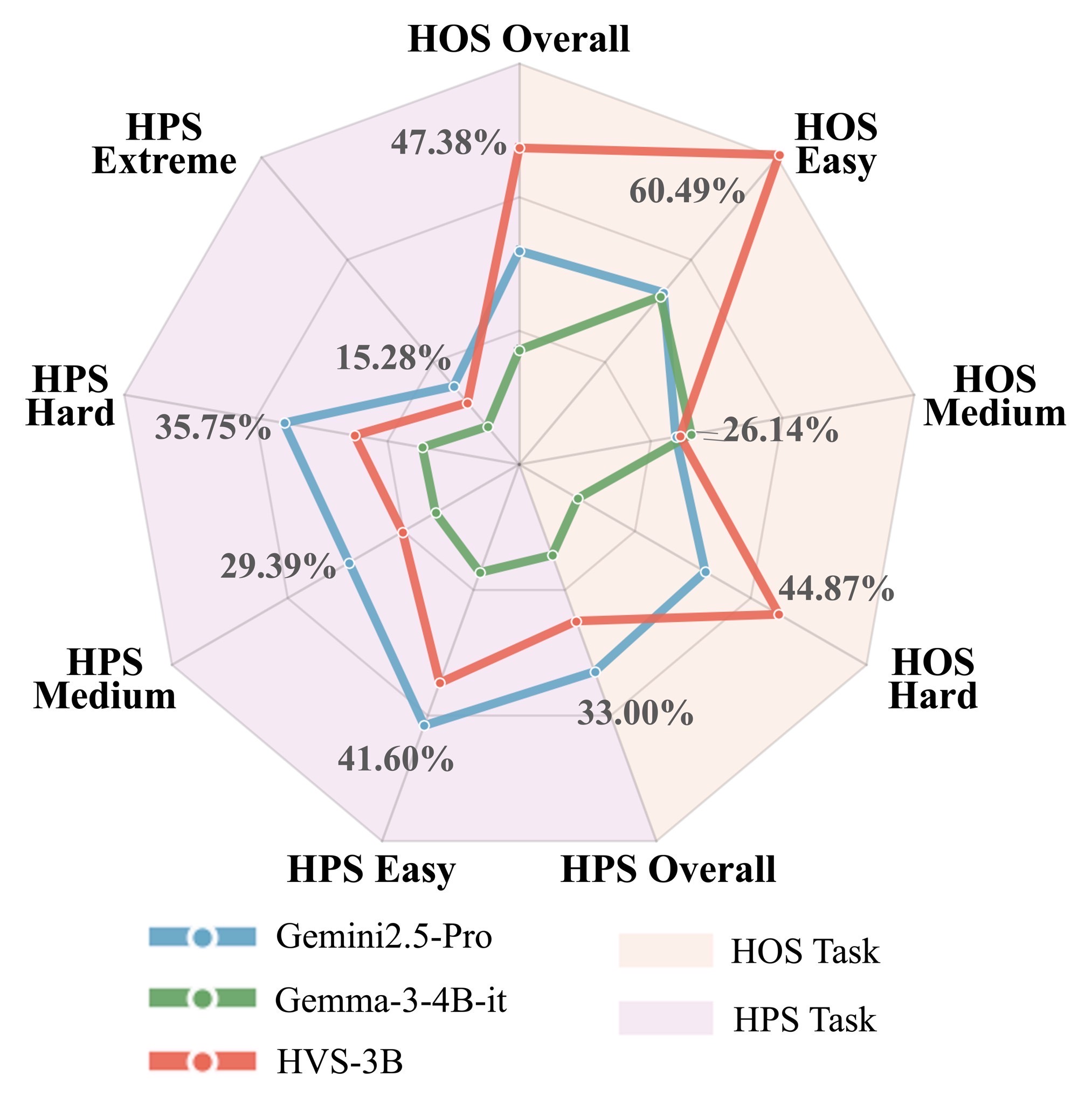}
    % \vspace{0.25em}\footnotesize Overview of split/statistics.
  \end{minipage}

  % \vspace{-0.6em}
  \vspace{-1mm}
\end{table*}

\begin{figure*}[t]
    \begin{center}
        \includegraphics[width=1\textwidth]{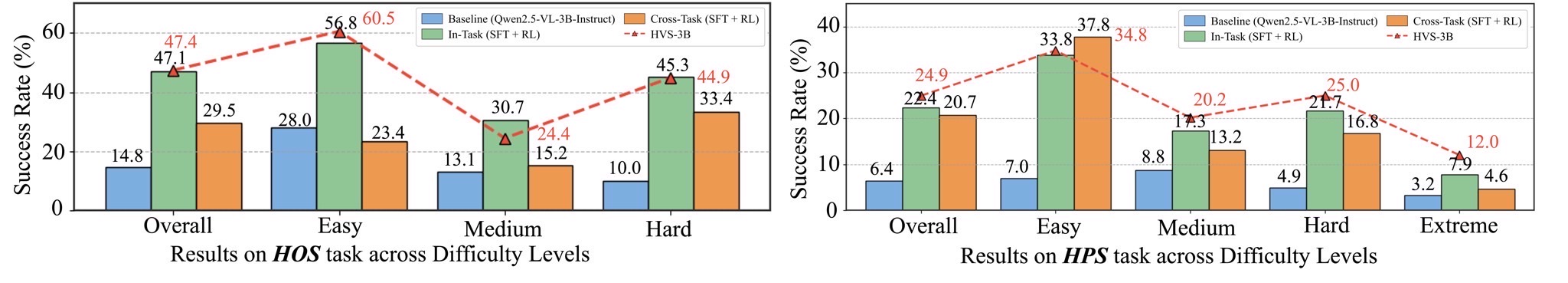}
        \vspace{-7mm}
        \caption{Comparison of \textbf{In-task} (train and test on the same task family) and \textbf{Cross-task} (train on one task family and test on the other).}
        \label{fig:task_performance.png}
        \vspace{-6mm}
    \end{center}
\end{figure*}
% \vspace{-2mm}
\subsection{Benchmark Construction}\label{subsec:bench:bench}

\noindent\textbf{Task Annotation.} We annotate each panoramic scene in a perspective-view interface that renders narrow–field-of-view (FOV) images from the panorama at known viewing angles $(\phi, \gamma)$. Annotators freely rotate the virtual camera to inspect the scene, identify a suitable embodied search task, write a natural-language instruction, and mark the target by drawing a tight bounding box that specifies its optimal direction. The bounding box is then back-projected onto the panorama, and its center yields the optimal target direction $(\phi^*, \gamma^*)$. For \hps, we retain only $(\phi^*)$ as the environment can be well approximated by a planar ground geometry.

\noindent\textbf{Cold-Start Data Curation.} To construct high-quality multi-turn trajectories for SFT, we select a subset of annotated task instances and augment them with structured chain-of-thought (CoT) rationales by prompting a strong MLLM (GPT-4o~\cite{gpt4o_blog}). For each annotation step, given the task instruction, current observation, and the human-provided optimal action (rotation angle or submit), GPT-4o is prompted to produce a concise, observation-grounded rationale explaining why the action is appropriate in context. We employ a \textit{human-in-the-loop} protocol in which annotators review, and refine the generated rationales to eliminate hallucinations, ensure grounding in visible scene evidence, and enforce stylistic consistency. The resulting dataset consists of \textbf{2,000} multi-turn trajectories containing visual observations, verified CoT rationales, and actions, which we use to bootstrap SFT. In total, six annotators dedicated 250 hours to embodied question annotation and CoT refinement.

\noindent\textbf{Difficulty Taxonomy.} For \hos, we define task difficulty based on the initial visibility of the target object. We calculate a visibility ratio $d$ as the fraction of the object area visible in the initial viewpoint compared to the complete area of the object. Higher visibility offers stronger perceptual cues and reduces exploratory burden, whereas lower visibility demands visual exploration during search. We therefore categorize \hos~samples into \textit{Easy}, \textit{Medium}, and \textit{Hard} (see \cref{fig:combined_chart.png} Bottom Right; visualizations in~\cref{fig:HOS_example}). For \hps, difficulty depends on whether the scene contains textual cues and whether the visual or textual cues align with the actual path direction. These two factors jointly define four difficulty levels (see~\cref{fig:combined_chart.png} Bottom Right; visualizations in~\cref{fig:HPS_easy}-\cref{fig:HPS_extreme}).

\section{Experiment}\label{sec:exp}

\subsection{Experiment Setup}\label{subsec:exp:setup}
\noindent\textbf{Implementation Details.} (1)~\textit{Training Setup}: We finetune the model on a mixed object and path search dataset. The SFT training environment is implemented with LLaMA-Factory~\citep{zheng2024llamafactory}, and the RL training is built on an open-source framework VAGEN~\citep{wang2025vagen}. The full training details can be found in Appendix~\ref{hp} for more details. We perform full-parameter SFT on Qwen2.5-VL-3B-Instruct for 3 epochs; the resulting model is HVS-3B (w/ SFT only). For the RL stage, we train with RL for 70 steps. We denote the obtained model as HVS-3B. The prompts used in our experiments are provided in Appendix~\cref{prompt}. (2)~\textit{Benchmark Setup}: We evaluate open-source and proprietary models that support multi-image input. We use the same evaluation environment as RL training. See Appendix for the detailed training-test split for the proposed two tasks. 
\noindent\textbf{Evaluation Metric.}
A trial is evaluated as success if the submitted final viewing direction $(\hat{\phi},\hat{\gamma})$ falls inside a bbox-centered tolerance region $\left[\,\phi^{*}-\tau_{\phi},\, \phi^{*}+\tau_{\phi}\,\right]
\times\left[\,\gamma^{*}-\tau_{\gamma},\, \gamma^{*}+\tau_{\gamma}\,\right]
$ around the objective direction $(\phi^*,\gamma^*)$ (the spherical direction of annotated bbox center). The parameters of tolerance are defined as: \(\tau_\phi=\max\!\big(\tfrac{w_\phi}{2},\,\tau_{\phi}\)) and \(\tau_\gamma=\max\!\big(\tfrac{w_\gamma}{2},\,\tau_{\gamma}\)), where $w_\phi$ and $w_\gamma$ are the angular width and height of the bounding box. 
Following Sec.~\ref{hvs}, $(\hat{\phi},\hat{\gamma})$ are evaluated for \textbf{\textit{HOS}} task, whereas for \textbf{\textit{HOS}} we assess only$(\hat{\phi})$.
The tolerances are $\tau_{\phi}=30^\circ,\tau_{\gamma}=20^\circ$  for \hos\ (to mimic the human foveation) and $\tau_{\phi}=10^\circ$ for \hps\ (to reflect the requirement for precise motion direction). We report success rate (\%) by task and difficulty.

\begin{figure*}[t]
    \begin{center}
        \includegraphics[width=0.96\textwidth]{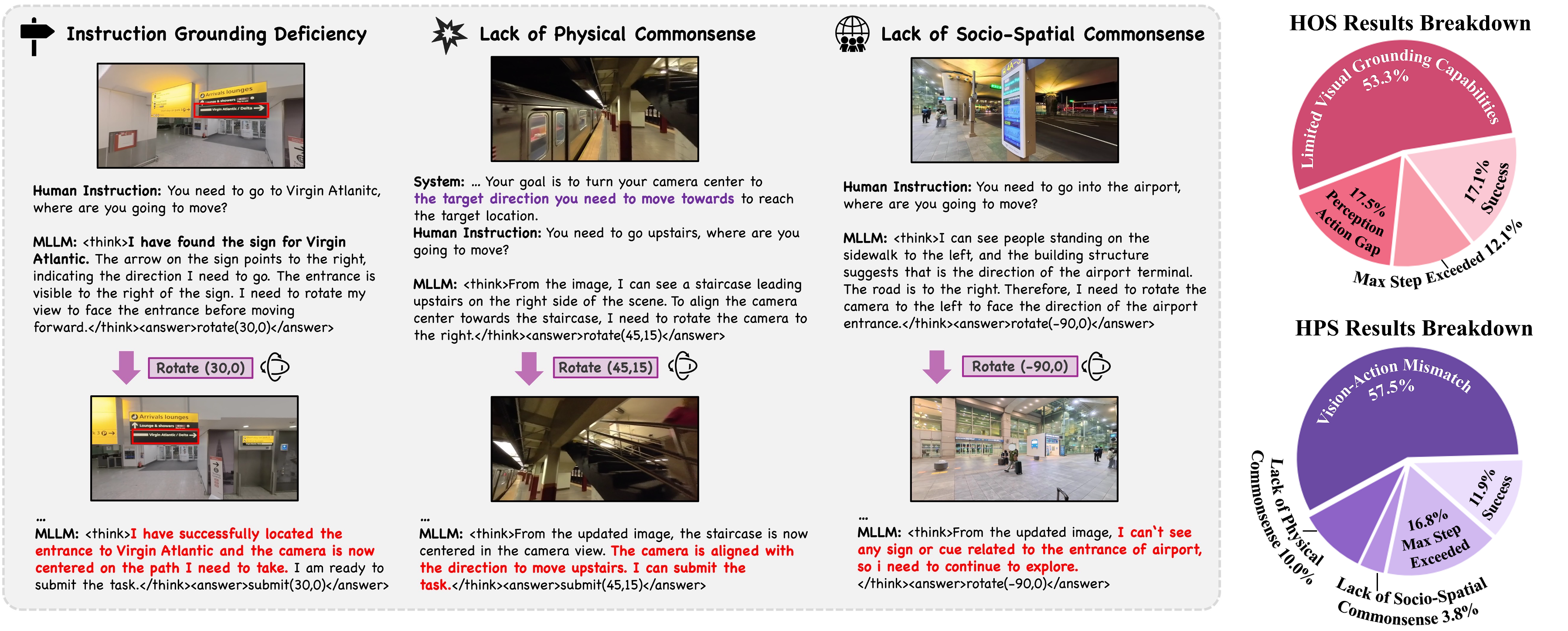}
        \vspace{-2mm}
        \caption{\textbf{Left: }\textbf{Failure cases in \hps.} (a) Vision-action mismatch. (b) Attempting to traverse an impassable surface instead of using an adjacent staircase. (c) Missing socio-spatial conventions (\eg, airport entrance cues), resulting in a fruitless search. \textbf{Right: }\hstar\ results breakdown of Gemma3-4B-it.}
        \vspace{-6mm}
        \label{fig:error-analysis}
    \end{center}
\end{figure*}

\subsection{Probing Embodied Visual Search in MLLMs}\label{subsec:exp:probing}

\noindent\textbf{Main Results.} \Cref{tab:bench} shows a large performance gap between proprietary and open-weight models, with Gemini2.5-Pro standing out as the strongest overall performer in \hos~(31.96) and \hps~(33.00). The Gemma-3 series achieves the best results among open-weight models. Interestingly, a larger model size does not guarantee better performance. For both the Gemma-3 and Qwen2.5-VL series, the smaller 4B/3B models surpass their larger 12B/7B counterparts in \hos, while performing on par in \hps.

\noindent\textbf{Error Analysis.} In \hos, errors arise from (1) \textbf{\textit{\textcolor{TakeawayRed}{limited visual grounding capabilities}}} (\eg, the agent fails to reliably identify targets in cluttered environments), and (2) \textbf{\textit{\textcolor{TakeawayGreen}{perception-action gap}}} (\eg, agent detects a target but cannot perform fine-grained foveation). In \hps, we identify three types of errors: (1) \textit{\textbf{\textcolor{TakeawayBlue}{vision-action mismatch}}}: The model perceives visual cues (\eg, signs) but fails to translate them into physical actions; (2) \textbf{\textit{\textcolor{Purple}{lack of physical commonsense}}}: Actions violate 3D constraints (\eg, attempting to pass through walls); \textbf{\textit{(3) {\textcolor{orange}{lack of socio-spatial commonsense}}}}: The model misses implicit rules and norms of built environments (\eg, the function of stairs, police tape, and crosswalks), as shown in Fig.~\ref{fig:error-analysis} Left. The breakdown of results is shown in~\cref{fig:error-analysis} Right, and additional examples are provided in Appendix~\cref{error_example}. These findings suggest:

\begin{redbox}
% \faBookmark\quad 
\textit{MLLMs can form linguistically grounded spatial models for passive world description, but not physically grounded ones for embodied world interaction.}
\end{redbox}

\subsection{On the Role and Limits of Post-Training}\label{subsec:exp:post}

\noindent\textbf{Effectiveness of SFT and RL.} Our post-training framework demonstrates significant improvements over the base model. As shown in \cref{tab:bench}, SFT contributes the majority of performance gains: on \hos, SFT alone improves the overall score from 14.83 to 40.83 ($\uparrow$26.00), while on \hps, it elevates performance from 6.44 to 23.00 ($\uparrow$16.56). Subsequent RL provides additional but more modest gains: $\uparrow$6.55 on object search and $\uparrow$1.94 on path search. This indicates that \textit{{SFT establishes the fundamental task capability, while RL serves as a refinement step for further optimization}}. Specifically, we find that post-training improves several key abilities: (1) \textit{precise control over rotation angles}, (2) \textit{the use of large-angle turns to explore new areas}, and (3) \textit{the capacity to act on directional signs} (see Appendix~\ref{case_study} for case studies). In addition, applying RL directly without prior SFT degrades the instruction following capability of the models.

\noindent\textbf{Task-Dependent Efficacy.} The benefits of post-training vary by task complexity. For the simpler object search, our model (47.38) outperforms the state-of-the-art proprietary model Gemini2.5-Pro (31.96). Yet for the more complex path search, its absolute score (24.94) falls short of the art (33.00). This gap suggests that \textit{{post-training has limitations in enhancing higher-order spatial reasoning capabilities}}.

\noindent\textbf{Negative Impact of RL on Complex Tasks.} In \hps, RL reduces performance on \textit{medium} difficulty from 23.03 to 20.18 and on \textit{extreme} difficulty from 14.81 to 12.04. These scenarios are characterized by a misalignment between visual cues and the optimal path, a pattern our taxonomy identifies as rather challenging. We hypothesise that this degradation may stem from {reward hacking}, where the model learns to exploit the reward signal rather than genuinely improving its reasoning capability. {\textit{This highlights the challenge of designing reward functions that consistently align with true task objectives across all difficulty levels.}} 

\noindent\textbf{Key Takeaway.} The disparate impact of our post-training method on object versus path search is telling. While SFT+RL yields substantial gains in object search across all difficulty levels, its improvements on the more demanding path search are more modest, with RL even degrading performance on some hard cases. This pattern suggests that:

\begin{greenbox}
\textit{Post-training can improve visual grounding and exploration for object search, but struggles to impart physical, spatial, and social commonsense for path search, as these are often implicit, situational, and procedural.}
\end{greenbox}

\subsection{Dissecting Object and Path Search}\label{subsec:exp:decoupling}
\noindent\textbf{In-Task Superiority with an Exception.}
As shown in \cref{fig:task_performance.png}, in-task training delivers peak performance, with one exception: a model trained on object search achieves 37.8\% on the \textit{easy} \hps~split, surpassing both the baseline (7.0\%) and the dedicated in-task \hps~model (33.8\%). We hypothesize that these easy tasks reduce to simple object searches where clear visual cues define the path, allowing the powerful object-finding skills from \hos~to transfer effectively.

\noindent\textbf{Cross-Task Generalization.}
We observe a clear bidirectional synergy: training on object search boosts path search performance from 6.4\% to 20.7\%, while training on path search elevates object search from 14.8\% to 29.5\%. This is because skills acquired from learning path search, like active exploration and path reasoning, confer a direct performance advantage in object search, while the visual grounding honed in object search reciprocally benefits path search.

\noindent\textbf{Mixed-Data Training.}
Training on a mixed object and path search dataset yields the best overall performance. Yet this comes with a key challenge: performance gains are unevenly distributed, as improvements on certain splits can reduce performance on others. Balancing this trade-off is essential for developing generalist humanoid agents.

\begin{figure}[t]
    \centering
    \includegraphics[width=\linewidth]{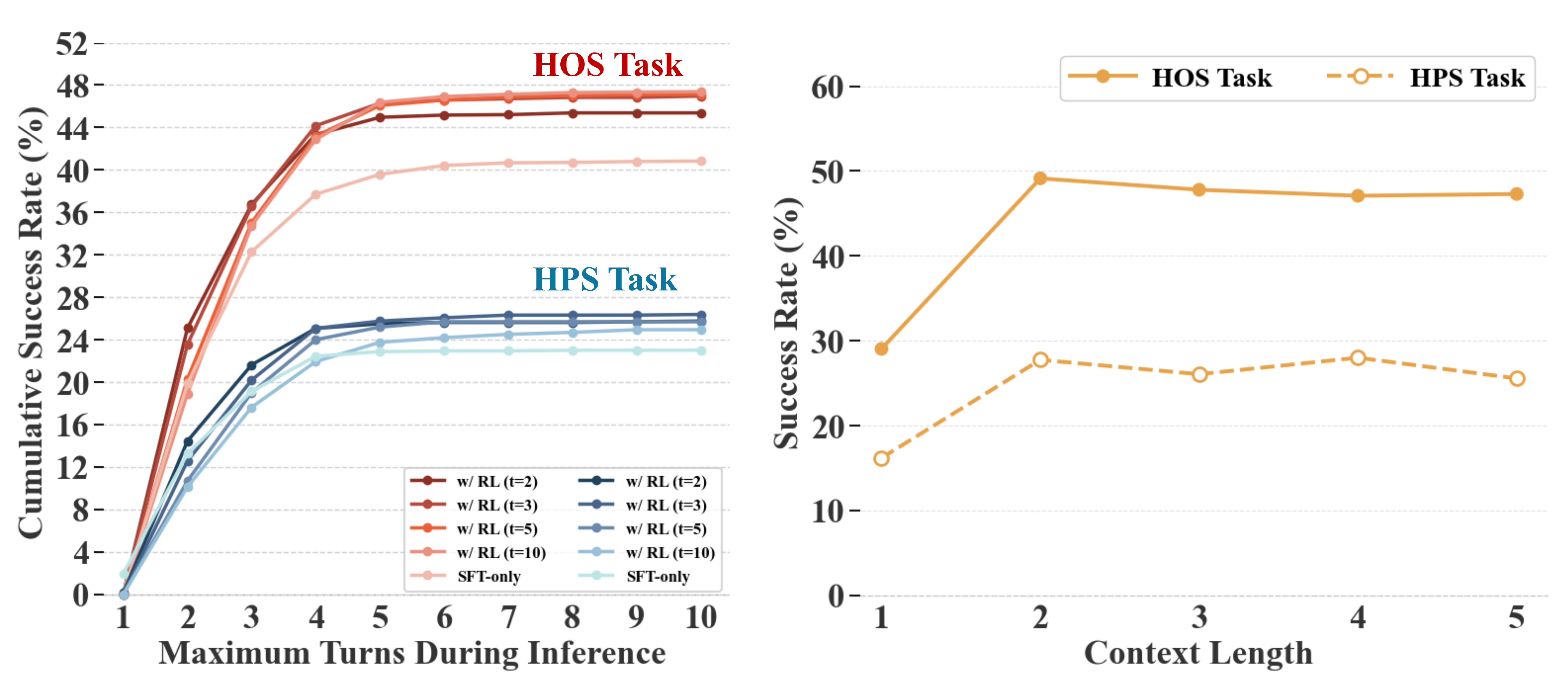}
    \vspace{-5mm}
    \caption{\textbf{Left:} Cumulative success rate by step before and after RL (t indicates maximum turn limit in RL training). \textbf{Right:} Impact of test-time context length on success rate.}
    \label{fig:tts}
    \vspace{-3mm}
\end{figure}

\begin{table}[t]
\centering
\resizebox{\columnwidth}{!}{
\begin{tabular}{lccccc}
\toprule
       & \multicolumn{5}{c}{\textbf{Humanoid Path Search}} \\
       % \cline{2-6}
       \textbf{Method} & \textbf{Overall} & \textbf{Easy} & \textbf{Medium} & \textbf{Hard} & \textbf{Extreme} \\
    \hline
    \rowcolor{tablegray}
    \textit{GRPO on \hps} & & & & & \\
    sft (baseline) &\cellcolor{blue!7} 23.44 & 26.00 &\cellcolor{blue!7} 24.56 & \cellcolor{blue!7}24.77 & \cellcolor{blue!7}12.50 \\
    form+corr  &22.38& 33.80 & 17.32 & 21.73 & 7.87\\
    form+corr+dist & 21.37 & \cellcolor{blue!7}34.40 & 15.13 & 20.09 & 6.94 \\
    form+dist  & 21.31 & 29.80 & 17.54 & 20.56 & 11.11 \\
    \bottomrule
\end{tabular}
}
\vspace{-2mm}
\caption{Results of GRPO with different reward shaping on \hps.}
\vspace{-4mm}
\label{tab:rw_shaping}
\end{table}

\subsection{Ablation Study}\label{subsec:exp:ablation}
\label{ablation}

\noindent \textbf{Reward Shaping.} We ablate three types of rewards for path search: (1) format + correctness, (2) format + correctness + distance-to-goal, and (3) format + distance-to-goal (see Appendix~\ref{reward_shaping_formulation} for reward functions). All variants improve performance only on the \textit{easy} split, often degrading harder levels (\cref{tab:rw_shaping}). This underlines the difficulty of path search and the need for more advanced learning algorithms.

\noindent \textbf{Training Rollout and Context Length.}
As \cref{fig:tts} shows, models trained with short GRPO rollouts can achieve satisfactory performance through test-time scaling, and match the performance of models trained with longer rollouts (10 turns) while converging faster. This ensure training efficiency without sacrificing final performance. Meanwhile, a short context length of 2 rounds is enough on \hvs.

\begin{figure}[t]
    \centering
    \includegraphics[width=\linewidth]{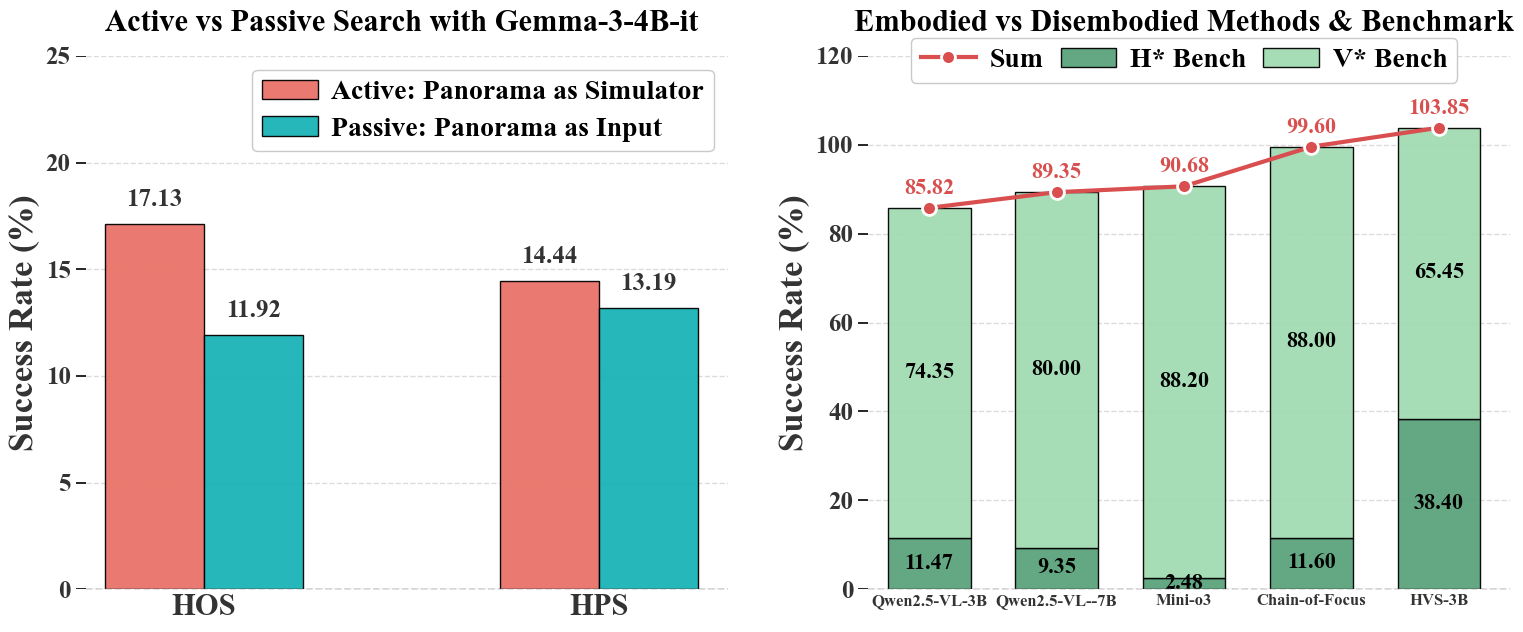}
    \vspace{-5mm}
    \caption{\textbf{Left:} Comparison of active and passive visual search. \textbf{Right:} Comparison of different visual search paradigms.}
    \vspace{-3mm}
    \label{fig:active}
\end{figure}

\noindent\textbf{Active vs. Passive.}
We compare active visual search, where an agent with a perspective view rotates to gather information, against the passive analysis of a complete panorama. The active paradigm is superior for two key reasons: (1) it mirrors efficient, human-like search strategies, and (2) it avoids panoramic distortions that conflict with MLLM training priors. Our empirical results validate this superiority; using Gemma-3-4B-it, we find the passive approach can degrade the performance (\cref{fig:active} Left). In addition, this emphasis on active spatial intelligence aligns our work with a growing body of research on active vision~\cite{pmlr-v305-kerr25a,yu2025egomi}.

\noindent\textbf{Embodied vs. Disembodied Bench.} As shown in \cref{fig:active} Right, 2D methods like Mini-o3~\cite{lai2025mini} and Chain-of-Focus~\cite{zhang2025chain} achieve near-saturation performance on the disembodied \textit{\textbf{V$^*$~Bench}} (88.2\% and 88.0\%, respectively), indicating that \textit{visual search within a static 2D image is no longer challenging for MLLMs}. However, their performance plummets on our embodied \hstar, with success rates dropping to a mere 2.5\% and 11.6\%. This stark contrast demonstrates that \textit{capabilities learned from passive Internet data do not transfer to embodied active interaction in 3D}. Actually, our HVS-3B model achieves a success rate of only 38.4\%, highlighting that \textit{\hvs~remains a wide-open research problem}. Notably, our model maintains a satisfactory 65.5\% success rate on \textit{\textbf{V$^*$~Bench}}. This suggests that: 
\begin{bluebox}
\textit{Our model learns 3D embodied search without compromising its 2D visual search ability too much, indicating a promising path toward a unified model capable of operating in both physical and digital realms.}
\end{bluebox}

\section{Discussion and Future Work}

We study MLLM-powered humanoid visual search in the wild by introducing the \hstar~and leveraging post-training to enhance the performance. Our analysis reveals that while post-training effectively improves low-level perceptual-motor abilities—such as visual grounding and exploration—it exposes fundamental bottlenecks in higher-level reasoning, which requires physical, spatial, and social commonsense. Furthermore, while RL boosts performance on simpler tasks, it can paradoxically degrade performance in complex scenarios. Future work should focus on designing more robust reward functions as well as more efficient vision tokenizers, developing pre-training methods that instill action-oriented spatial world knowledge, and balancing performance across task difficulties. Meanwhile, scaling up the collection of embodied search data is essential for fully unlocking visual-spatial reasoning in the wild.

\noindent\textbf{Acknowledgment.}
The work was supported in part through NSF grants 2514030, 2238968, 2345139, 2443404, and the NYU IT High Performance Computing resources, services, and staff expertise. S.X. acknowledges support from the MSIT ITP grant (RS-2024-00457882). The authors thank Haoxuan Wang for data collection, and Justin Kerr and Ken Goldberg for insightful discussion.

{
% \newpage
    \small
    \bibliographystyle{ieeenat_fullname}
    \bibliography{main}
}

\clearpage

\newcommand{\appendixroman}{
    \renewcommand{\thesection}{\Roman{section}}
    \renewcommand{\thesubsection}{\thesection\Roman{subsection}}
    \renewcommand{\thetable}{\Roman{table}}
    \renewcommand{\thefigure}{\Roman{figure}}
    \setcounter{section}{0}
    \setcounter{figure}{0}
    \setcounter{table}{0}
}

\section*{Appendix}
\appendixroman

\subsection{Geographical Distribution of \hstar}\label{geographical_distribution}

\begin{figure*}[t]
    \begin{center}
        \includegraphics[width=0.91\textwidth]{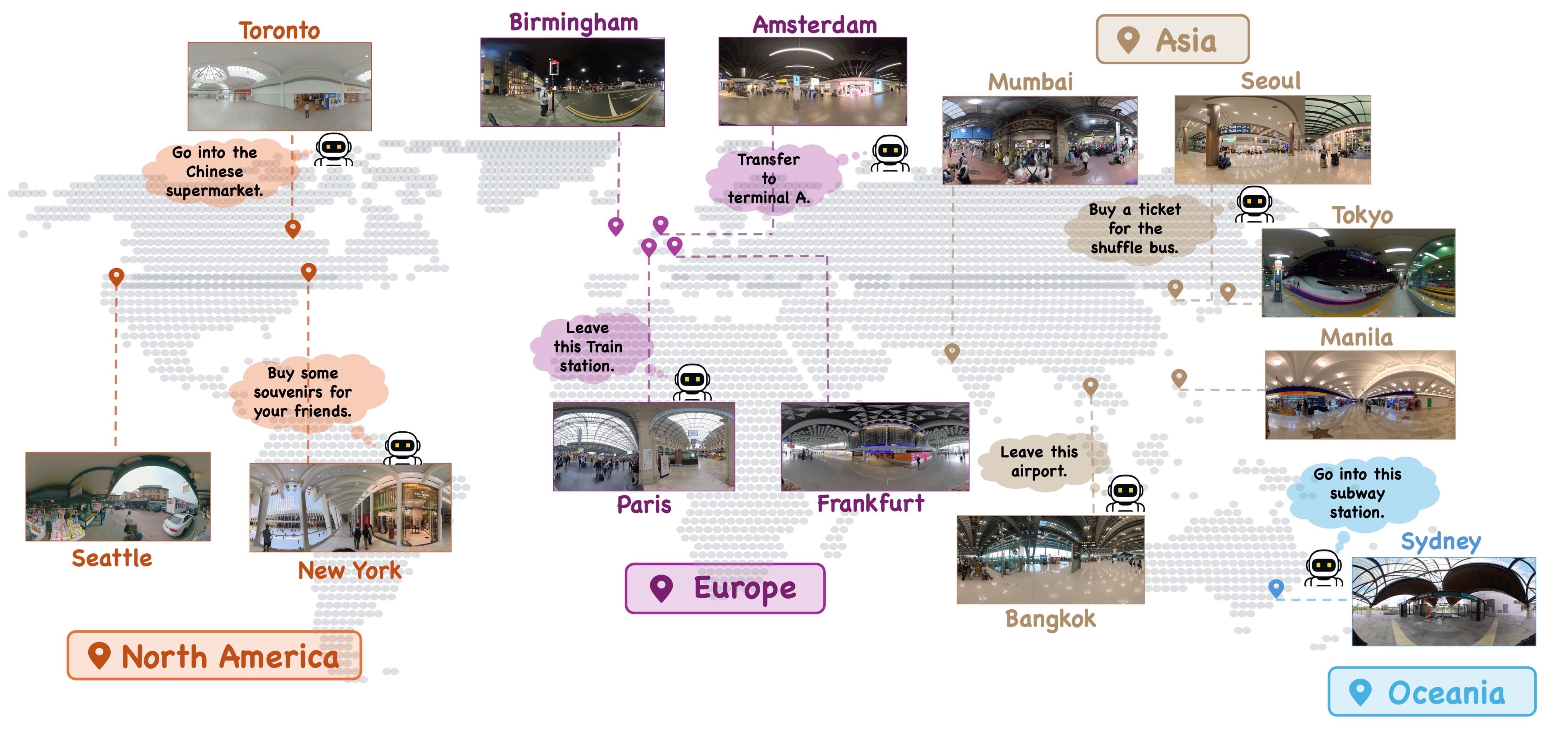}
        \caption{\hstar~aggregates panoramic videos from diverse global locations, featuring visually cluttered environments.}
        \label{fig:geographical}
    \end{center}
    \vspace{-5mm}
\end{figure*}
As shown in Fig.~\ref{fig:geographical}, our dataset exhibits broad geographical coverage, including thirteen cities across twelve countries and four continents. This diversity is reflected in the wide range of architectural styles, languages and scripts on signage, and environmental conditions.

\subsection{HVS Task Difficulty Visualization}
\subsubsection{HOS Difficulty Visualization}\label{hos_example}
\begin{figure*}[t]
    \begin{center}
        \includegraphics[width=\textwidth]{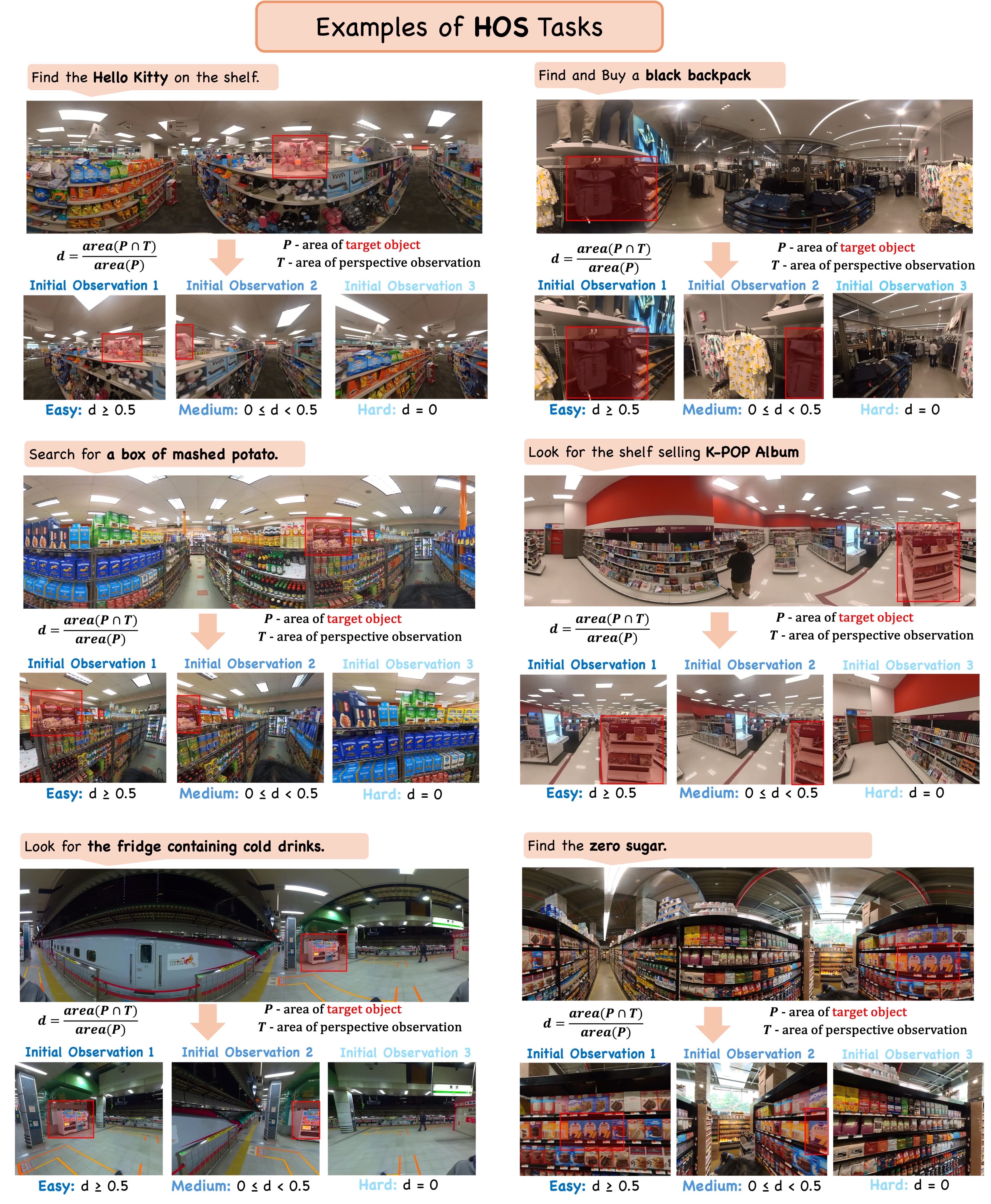}
        \caption{Visualizations of \textbf{\textit{HOS}} task instances.}
        \label{fig:HOS_example}
    \end{center}
\end{figure*}

\Cref{fig:HOS_example} illustrates the three difficulty levels of our \textbf{\textit{HOS}} task with concrete examples. For each keyframe, we overlay the target object's area on the initial view and provide its visibility ratio $d$. From left to right, we show one example per level: \textbf{Easy} (mostly visible), \textbf{Medium} (partially visible), and \textbf{Hard} (invisible), demonstrating the model's initial observation.

\subsubsection{HPS Difficulty Visualization}\label{hps_example}
\begin{figure*}[t]
    \begin{center}
        \includegraphics[width=\textwidth]{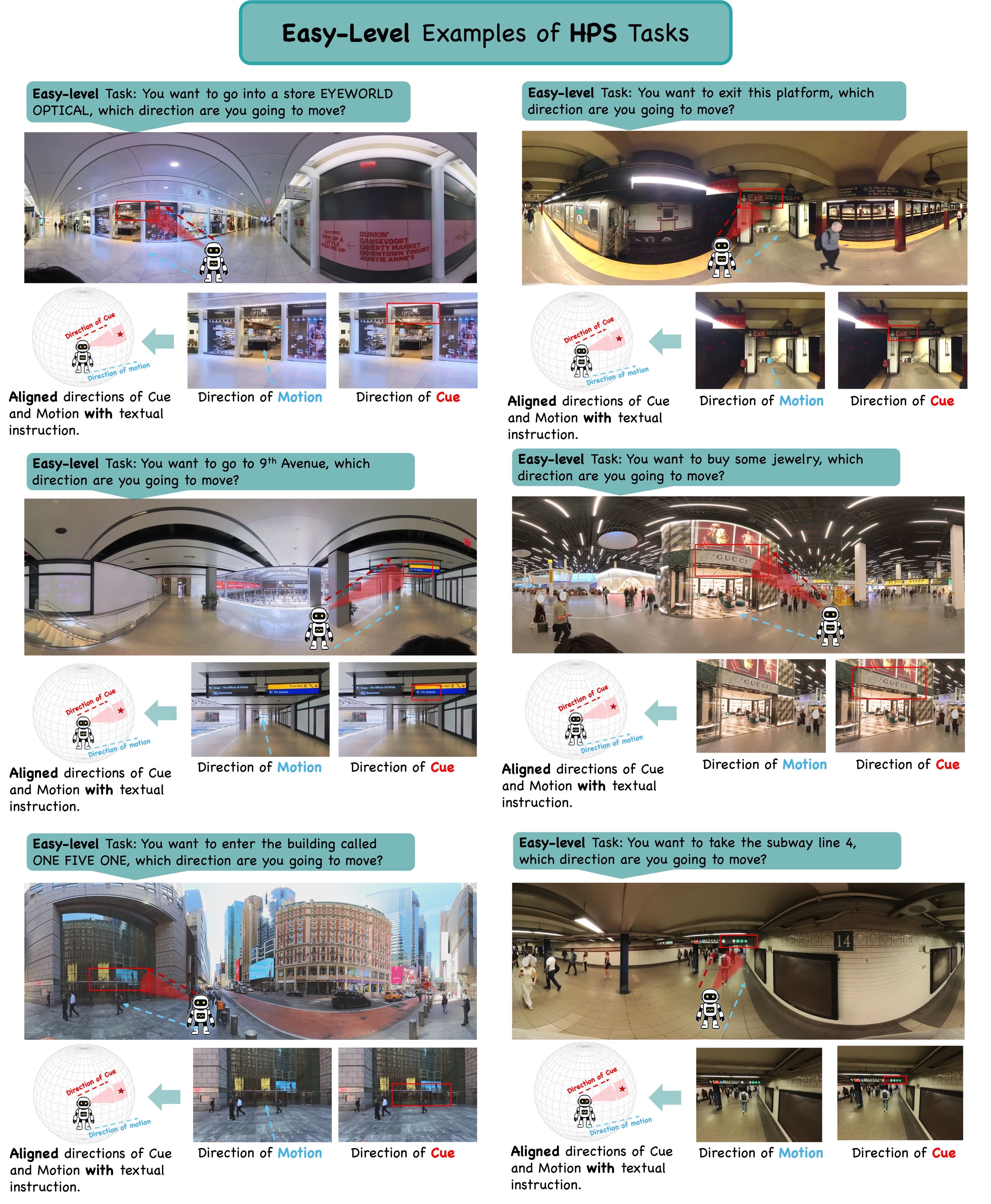}
        \caption{Visualizations of easy-level \textbf{\textit{HPS}} task instances.}
        \label{fig:HPS_easy}
    \end{center}
\end{figure*}

\begin{figure*}[t]
    \begin{center}
        \includegraphics[width=\textwidth]{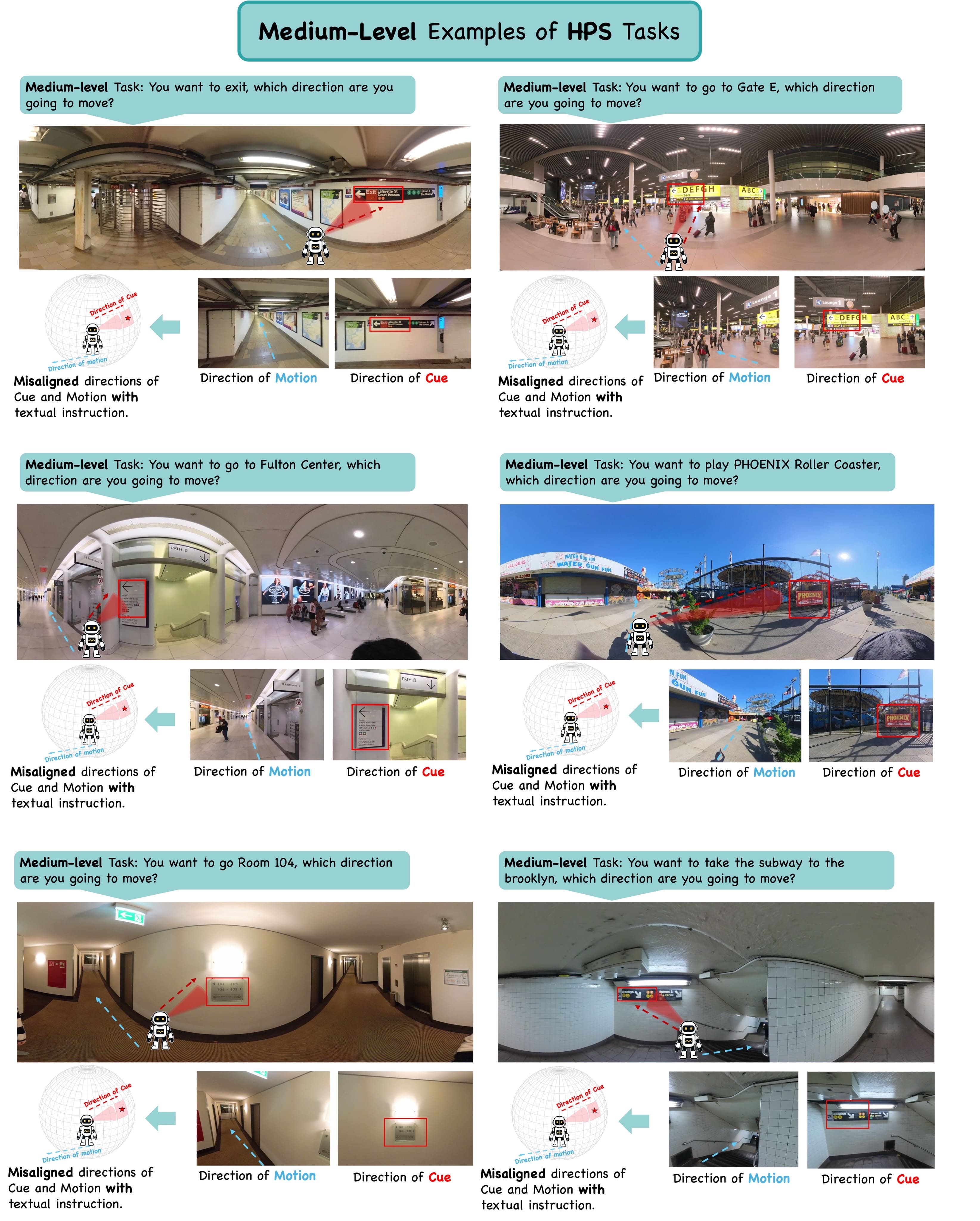}
        \caption{Visualizations of medium-level \textbf{\textit{HPS}} task instances.}
        \label{fig:HPS_medium}
    \end{center}
\end{figure*}

\begin{figure*}[t]
    \begin{center}
        \includegraphics[width=\textwidth]{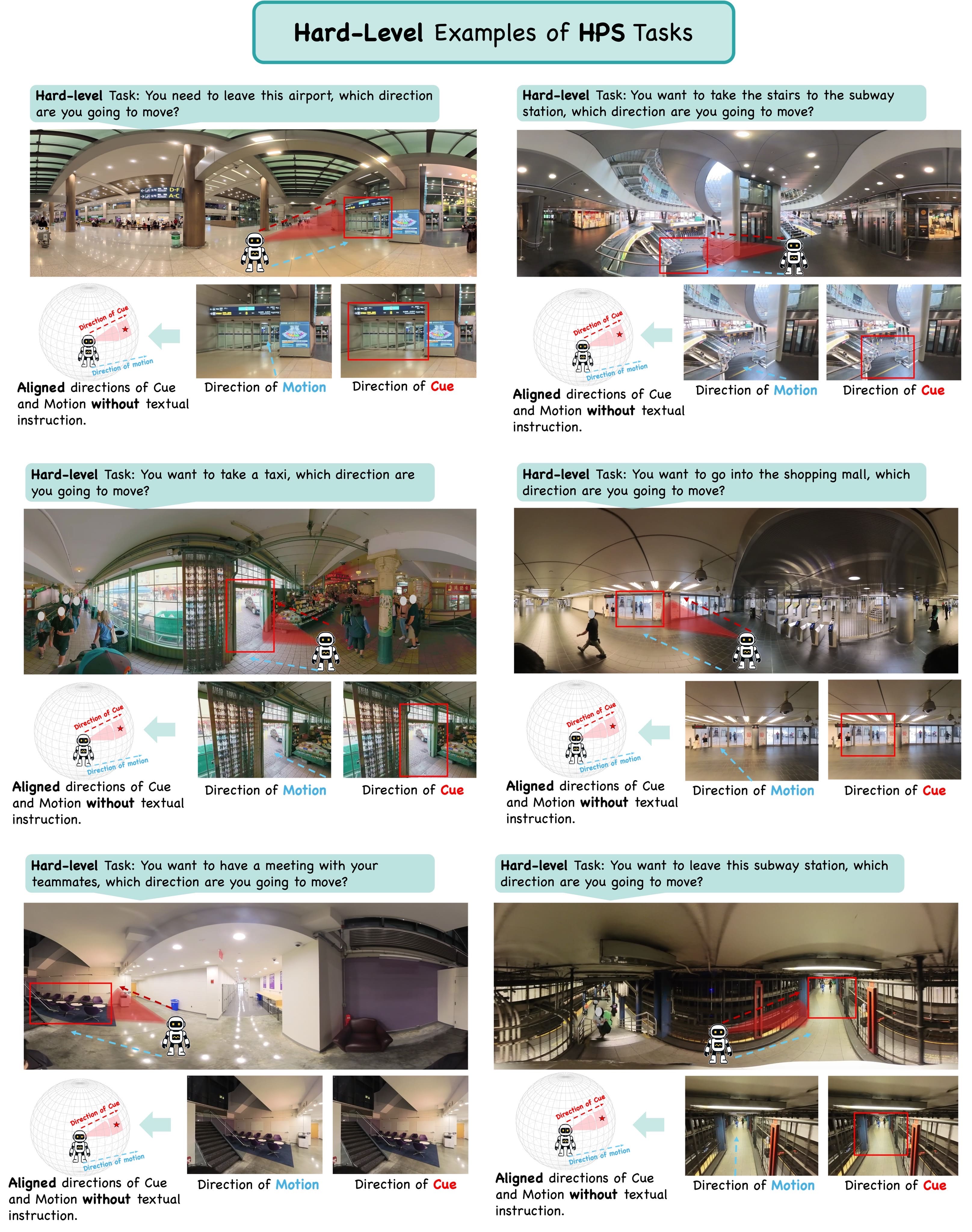}
        \caption{Visualizations of hard-level \textbf{\textit{HPS}} task instances.}
        \label{fig:HPS_hard}
    \end{center}
\end{figure*}

\begin{figure*}[t]
    \begin{center}
        \includegraphics[width=\textwidth]{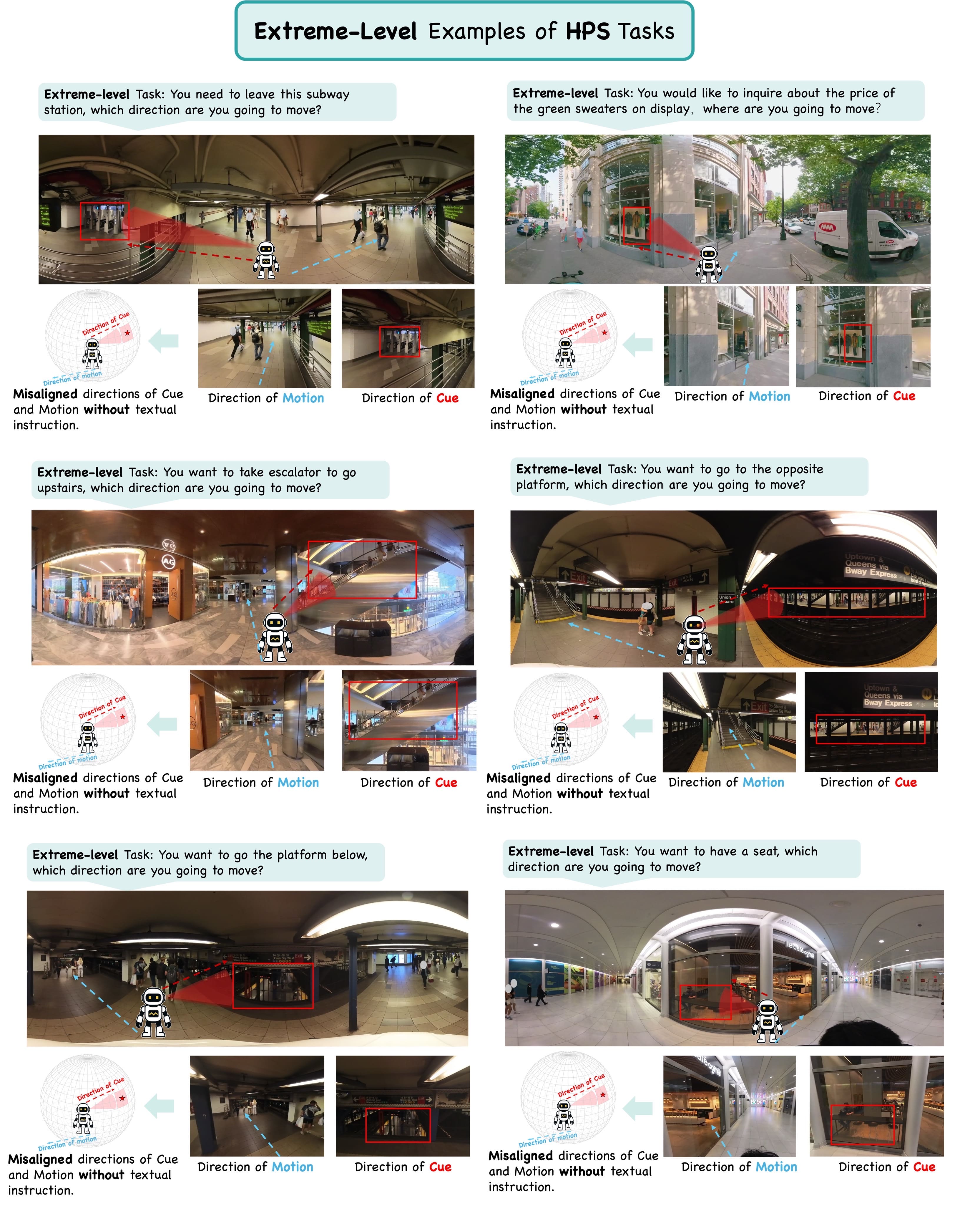}
        \caption{Visualizations of extreme-level \textbf{\textit{HPS}} task instances.}
        \label{fig:HPS_extreme}
    \end{center}
\end{figure*}

We define the four difficulty levels of the \textbf{\textit{HPS}} task based on two criteria: whether the relevant cue aligns with the navigable path and whether textual information is provided. Examples are shown in~\cref{fig:HPS_easy}-\cref{fig:HPS_extreme}.
 
\subsection{Training and Inference Prompts}
The natural-language prompts used for training and inference of the \textbf{\textit{HOS}} and \textbf{\textit{HPS}} tasks are shown in~\cref{prompt}.

\begin{figure*}[t]
\centering
\begin{tcolorbox}[
    colback=gray!10,
    colframe=black,
    width=\textwidth,
    title={\textbf{Inference Prompt}},
    sharp corners,
    boxrule=0.5pt
]
\# SYSTEM PROMPT - HOS \\
You are a robot and perform object searching tasks according to instructions. Your goal is to rotate the camera to center the target object in the camera view, and then submit the task. The camera center is presented as a green cross in the picture.\\

\# SYSTEM PROMPT - HPS \\
You are a robot and perform navigation tasks according to instructions. Your goal is to turn your camera center to the target direction you need to move towards to reach the target location. The camera center is presented as a green cross in the picture. Don't move in the unavailable direction, such as obstacles or gaps.\\

Actions you can take: rotate(yaw:int,pitch:int), submit(yaw:int,pitch:int)\\
rotate(yaw:int,pitch:int): rotate the camera in the yaw and pitch direction relative to the current direction. Yaw is the rotation angle in the x-y plane, pitch is the rotation angle in the y-z plane. Yaw angle $<$ 0 means rotate to the right, yaw angle $<$ 0 means rotate to the left. Pitch angle $>$ 0 means look up, pitch angle $<$ 0 means look down.\\
submit(yaw:int,pitch:int): submit the task with the current camera view with the target object at the center, yaw and pitch are the angles of the current camera view, which is reported by the environment.
You can only take one action at a time. The instruction will be provided with each observation. Look at the image carefully to complete the instruction.\\

\# FEW SHOT EXAMPLE (disabled after training)\\
Example:\\
Round 1:
image\_1\\
$<$think$>$I need to find the coffee machine. I can see a table on on my left, a couch in front of me, and a door to the right. The coffee machine is likely on the table, which is to my left.$<$/think$>$$<$answer$>$rotate(-45,0)$<$/answer$>$\\
Round 2:
Env\_feedback: Last action is executed successfully, your current direction (yaw,pitch) is (315,0).\\
image\_2\\
$<$think$>$From the secene, I see that by turning left 45 degrees, a kitchen table is in front of me. The coffee machine is on the left of the table and slightly lower than the camera center. I need to turn leftward and downward a little bit.$<$/think$>$
$<$answer$>$rotate(-30,-5)$<$/answer$>$\\
Round 3:
Env\_feedback: Last action is executed successfully, your current direction (yaw,pitch) is (285,-5).\\
image\_3\\
$<$think$>$The coffee machine is right now at the center of my camera, I think I can submit the task.$<$/think$>$
$<$answer$>$submit(285,-5)$<$/answer$>$\\
Round 4:
Env\_feedback: Success\\
You can take 1 action(s) at a time.
You should first give your thought process, and then your answer.\\
Your response should be in the format of:
$<$think$>$...$<$/think$>$$<$answer$>$...$<$/answer$>$\\
e.g. $<$think$>$I need to find the coffee machine. I can see a table on on my left, a couch in front of me, and a door to the right. The coffee machine is likely on the table, which is to my left.$<$/think$>$$<$answer$>$rotate(-45,0)$<$/answer$>$\\

\# USER PROMPT\\
After your answer, the extracted valid action is \{valid\_action\}.\\
The environment feedback is: \{env\_feedback\}\\
done: \{done\}\\
After that, the observation is:
\{observation\}\\
Human Instruction: \{instruction\}\\
Decide your next action.\\
You can take 1 action(s) at a time. You should first give your thought process, and then your answer.\\
Your response should be in the format of:
$<$think$>$...$<$/think$>$$<$answer$>$...$<$/answer$>$\\
\end{tcolorbox}
\caption{Prompt used for inference and rollout.}
\label{prompt}
\end{figure*}

\subsection{Objective Functions}
\label{objective func}
\paragraph{SFT stage.}
The SFT objective function is the expected negative log-likelihood (cross-entropy loss) over the dataset 
$\mathcal{D}^{SFT}$ which consists of task input $x$ and labeled trajectory $\mathcal{H}_T$:
\[
\min_{\theta} \; \mathbb{E}_{(x, \mathcal{H}_T) \sim \mathcal{D}^{SFT}} 
\left[ - \sum_{i=0}^{T-1} \log \pi_\theta(y_i, a_i \mid o_i, x, \mathcal{H}_i) \right].
\]

\paragraph{RL stage.} For each task, we sample $G$ times and get outputs $\{\omega_1,\omega_2,\dots,\omega_G\}$ where $\omega_i$ includes all the output tokens in the output sequence $\{y_0,a_0,y_1,a_1,\dots,y_{T-1},a_{T-1}\}$, then calculate the GRPO advantage to update the parameters.
The GRPO objective function is:

$$\begin{aligned}
&\mathcal{J}_{\text{GRPO}}(\theta) =\mathbb{E}{[
    (s_o,x,y) \sim \mathcal{D}^{RL},
    \{\omega_i\}_{i=1}^{G} \sim \pi_{\theta_{\text{old}}}(\Omega|s_o,x)
]}
    \\&  \frac{1}{G} \sum_{i=1}^G \frac{1}{|\omega_i|}\sum_{t=1}^{|\omega_i|}
    \\&  \bigg\{ 
    \min\bigg[
        \frac{\pi_\theta (\omega_{i,t}|s_o,x,\omega_{i,<t})}{\pi_{\theta_{\text{old}}} (\omega_{i,t}|s_o,x,\omega_{i,<t})}\hat{A}_{i,t}, 
        \mathrm{clip}(
            \frac{\pi_\theta (\omega_{i,t}|s_o,x,\omega_{i,<t})}{\pi_{\theta_{\text{old}}} (\omega_{i,t}|s_o,x,\omega_{i,<t})}\\
         &,
            1-\epsilon,
            1+\epsilon )\hat{A}_{i,t} 
        \bigg] 
     - \beta\,\mathbb{KL} ( \pi_\theta \Vert \pi_{\text{ref}} )
    \bigg\},
\end{aligned}
$$

where $\hat{A}_{i,t}$ denotes the group relative advantage at $\omega_{i,t}$:

$$
\hat{A}_{i,t} = \frac{r_i - \mathrm{mean} (r)}{\mathrm{std}(r)}.
$$

\subsection{Reward Shaping}
\label{reward_shaping_formulation}
We use rule-based reward function to calculate the reward of the trajectory, which includes correctness reward, format reward.
$$ r = r_{corr} + r_{form},$$
where:
$$
\begin{aligned}
r_{\text{corr}} &=
\begin{cases}
0.5, & \text{if the submitted answer satisfies the}\\
     & \quad\text{completion condition},\\
0,   & \text{otherwise},
\end{cases}
\\[6pt]
r_{\text{form}} &=
\begin{cases}
0.5, & \text{if the response is in $<$think$>$$<$/think$>$}\\
     & \quad\text{$<$answer$>$$<$answer$>$format},\\
0,   & \text{otherwise}.
\end{cases}
\end{aligned}
$$
Specially, we add a distance-to-goal reward for \hps. It is calculated by the distance of the final direction to the target bounding box.
$$
r_{dist} = \frac{\pi-d(\phi_T,\phi^*)+\pi-d(\gamma_T,\gamma^*)}{2\pi}.
$$
Distance to bounding box is calculated by:
$$
d(\alpha,\alpha^*) = |\alpha-(\alpha^*-\tau_\alpha)|+|\alpha-(\alpha^*+\tau_\alpha)|
$$
which remains a constant minimum value when the direction is in the bounding box.

\begin{figure*}[t]
    \begin{center}
        \includegraphics[width=\textwidth]{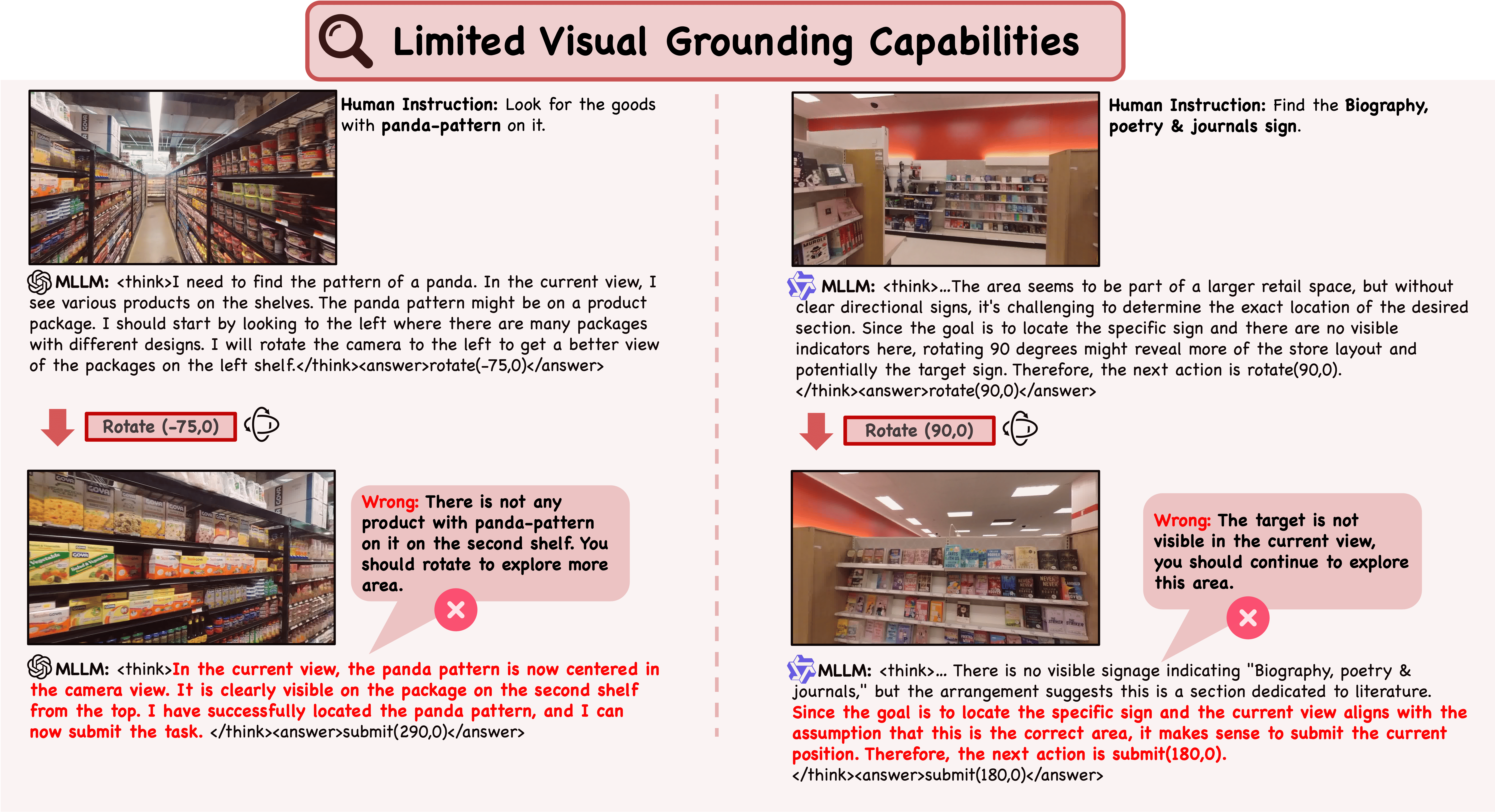}
        \caption{Qualitative Examples of \textbf{Limited Visual Grounding Capabilities} in \textbf{\textit{HOS}}.}
        \label{fig:HOS_error1}
    \end{center}
\end{figure*}

\begin{figure*}[t]
    \begin{center}
        \includegraphics[width=\textwidth]{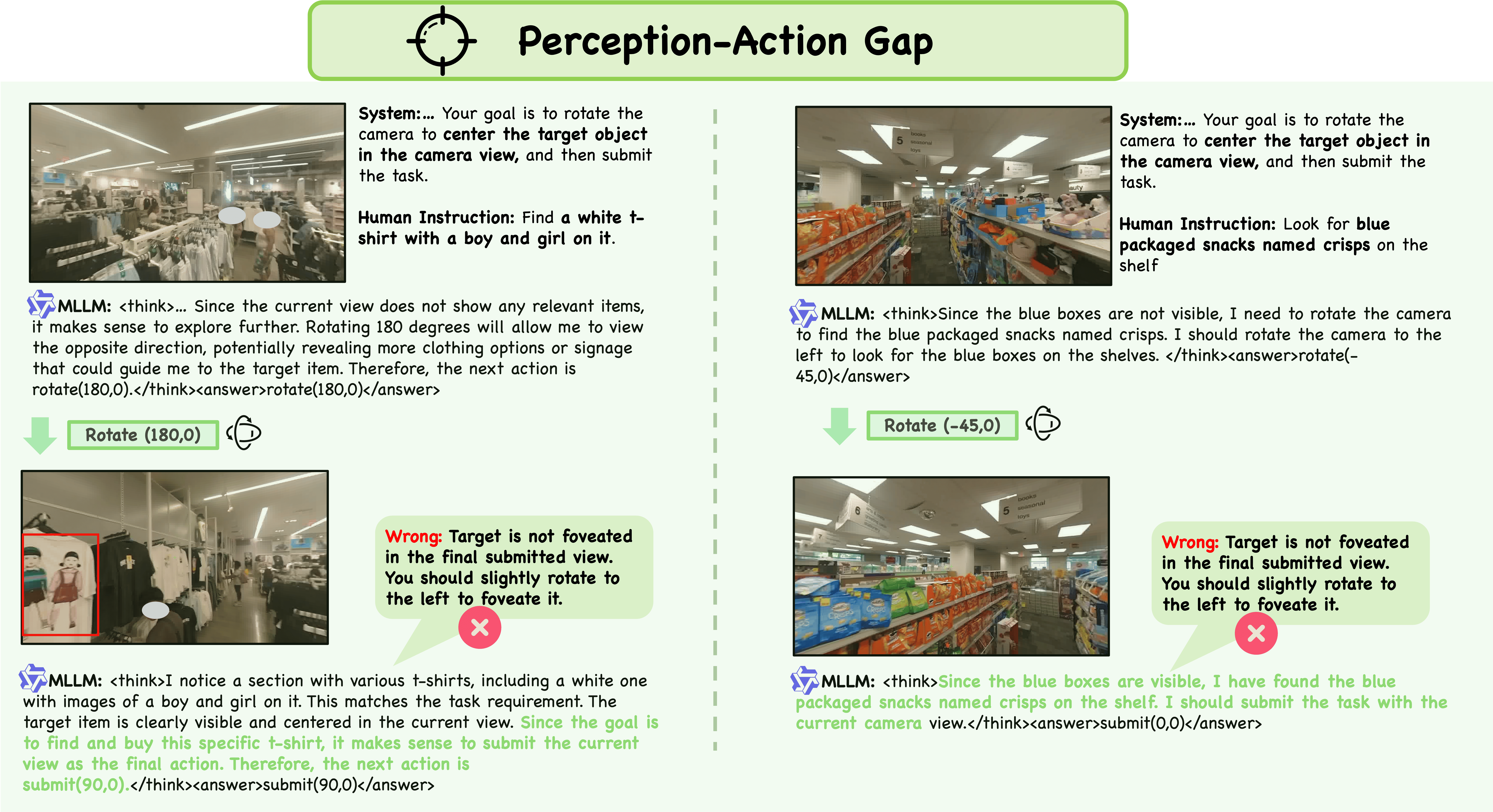}
        \caption{Qualitative Examples of \textbf{Perception-Action Gap} in \textbf{\textit{HOS}}.}
        \label{fig:HOS_error2}
    \end{center}
\end{figure*}

\begin{figure*}[t]
    \begin{center}
        \includegraphics[width=\textwidth]{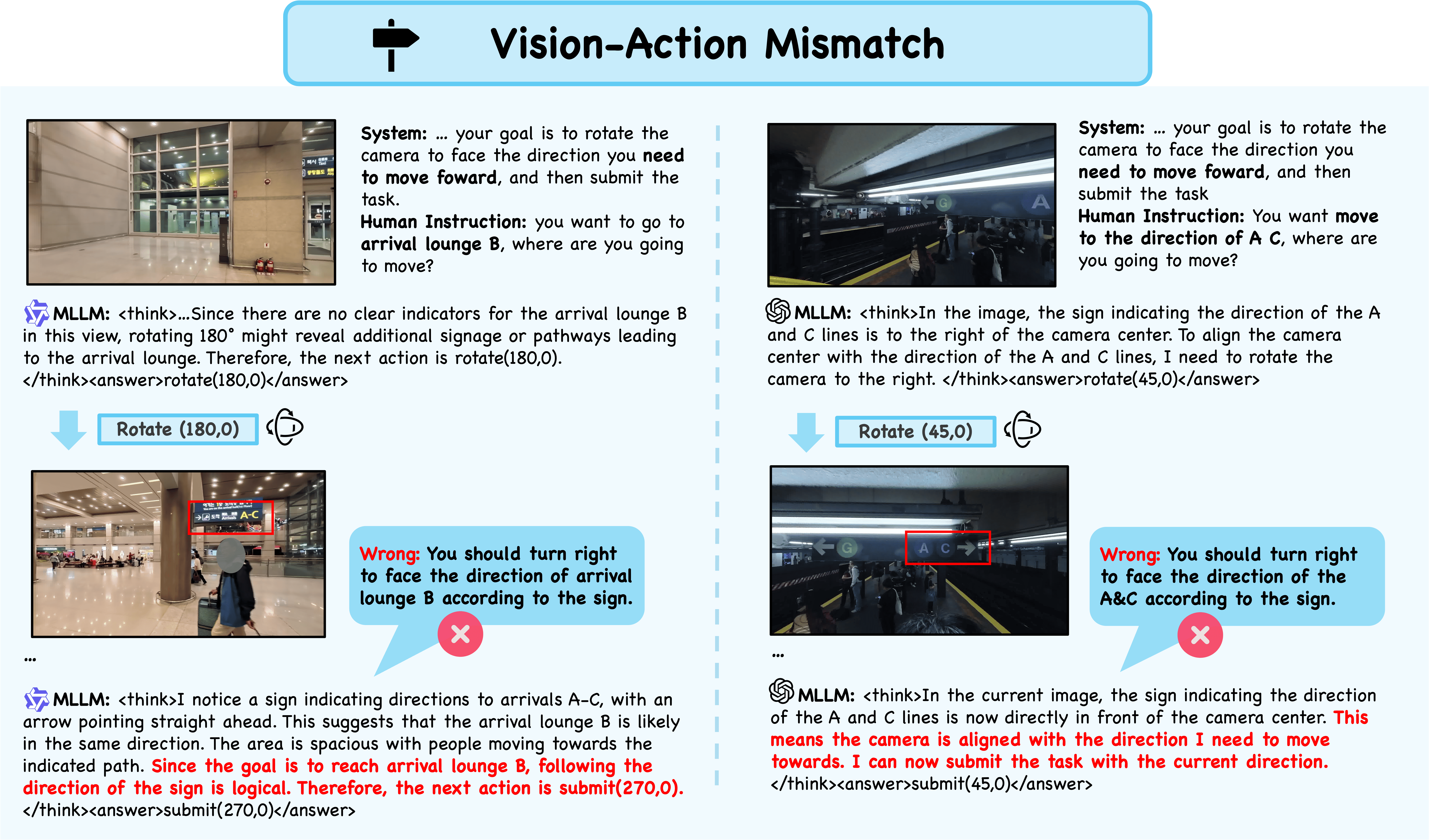}
        \caption{Qualitative Examples of \textbf{Vision-Action Mismatch} in \textbf{\textit{HPS}}.}
        \label{fig:HPS_error1}
    \end{center}
\end{figure*}

\begin{figure*}[t]
    \begin{center}
        \includegraphics[width=\textwidth]{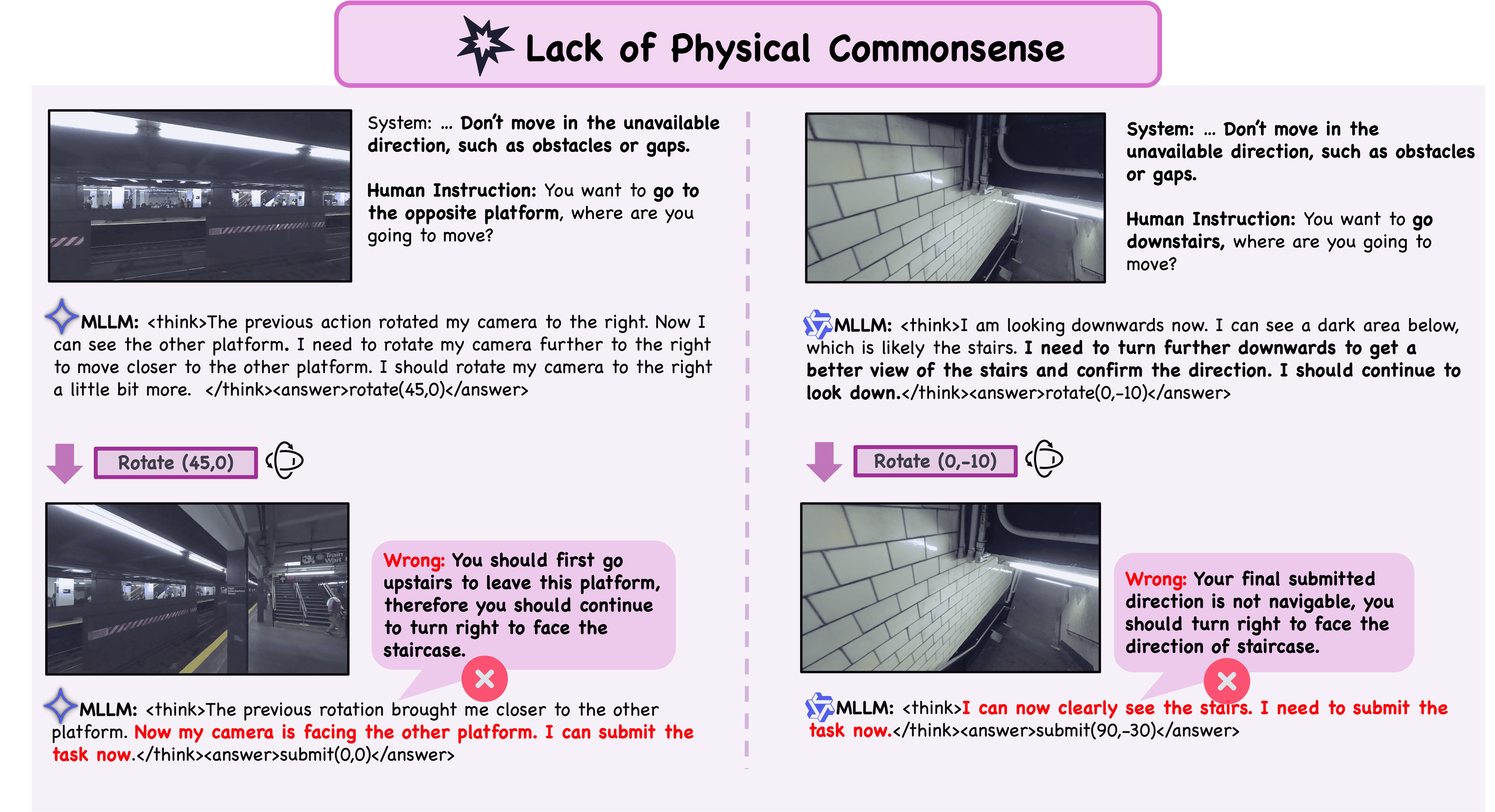}
        \caption{Qualitative Examples of \textbf{Lack
of Physical Commonsense} in \textbf{\textit{HPS}}.}
        \label{fig:HPS_error2}
    \end{center}
\end{figure*}

\begin{figure*}[t]
    \begin{center}
        \includegraphics[width=\textwidth]{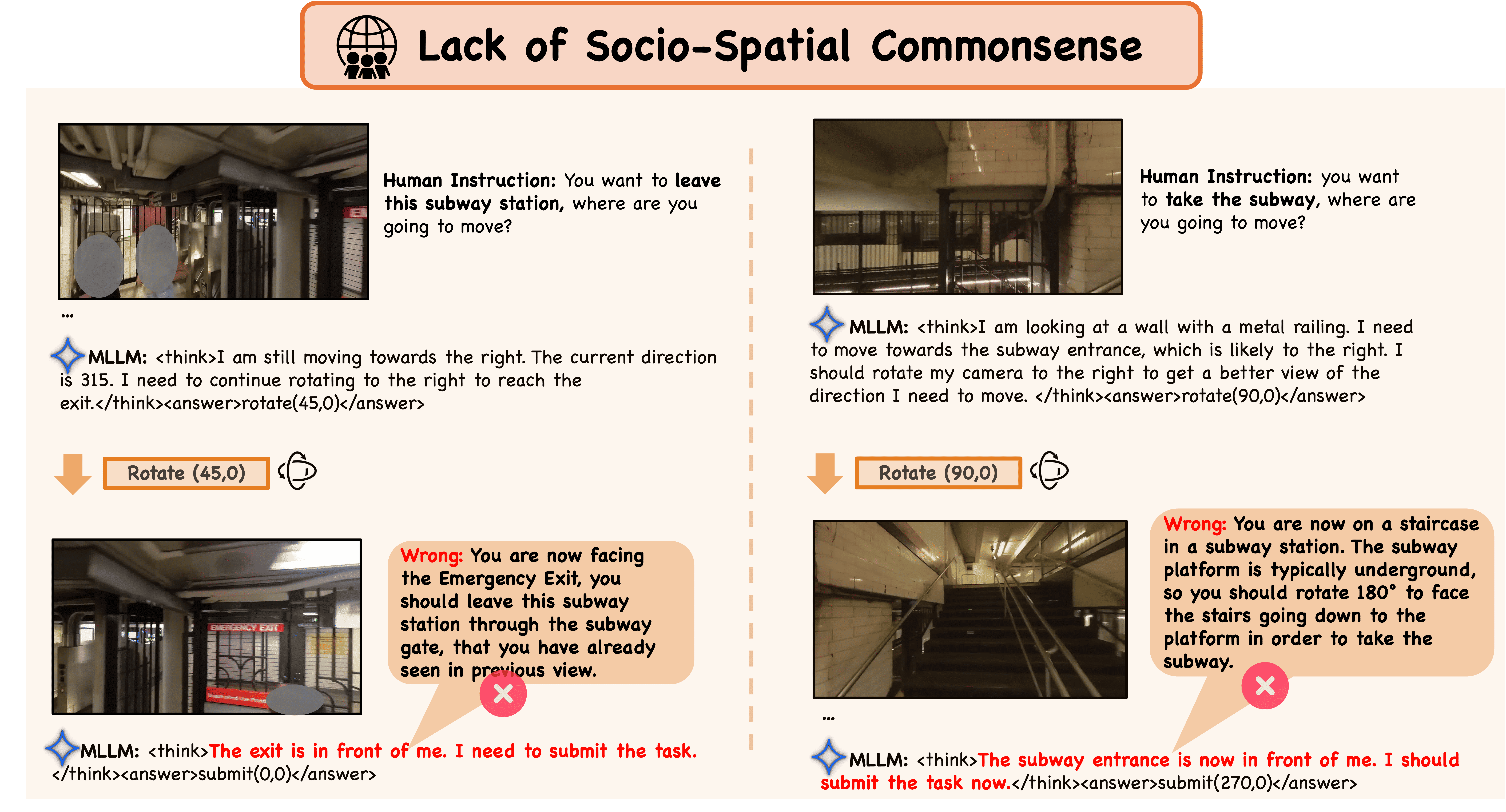}
        \caption{Qualitative Examples of \textbf{Lack
of Socio-Spatial Commonsense} in \textbf{\textit{HPS}}.}
        \label{fig:HPS_error3}
    \end{center}
\end{figure*}

\begin{figure*}[t]
    \begin{center}
        \includegraphics[width=\textwidth]{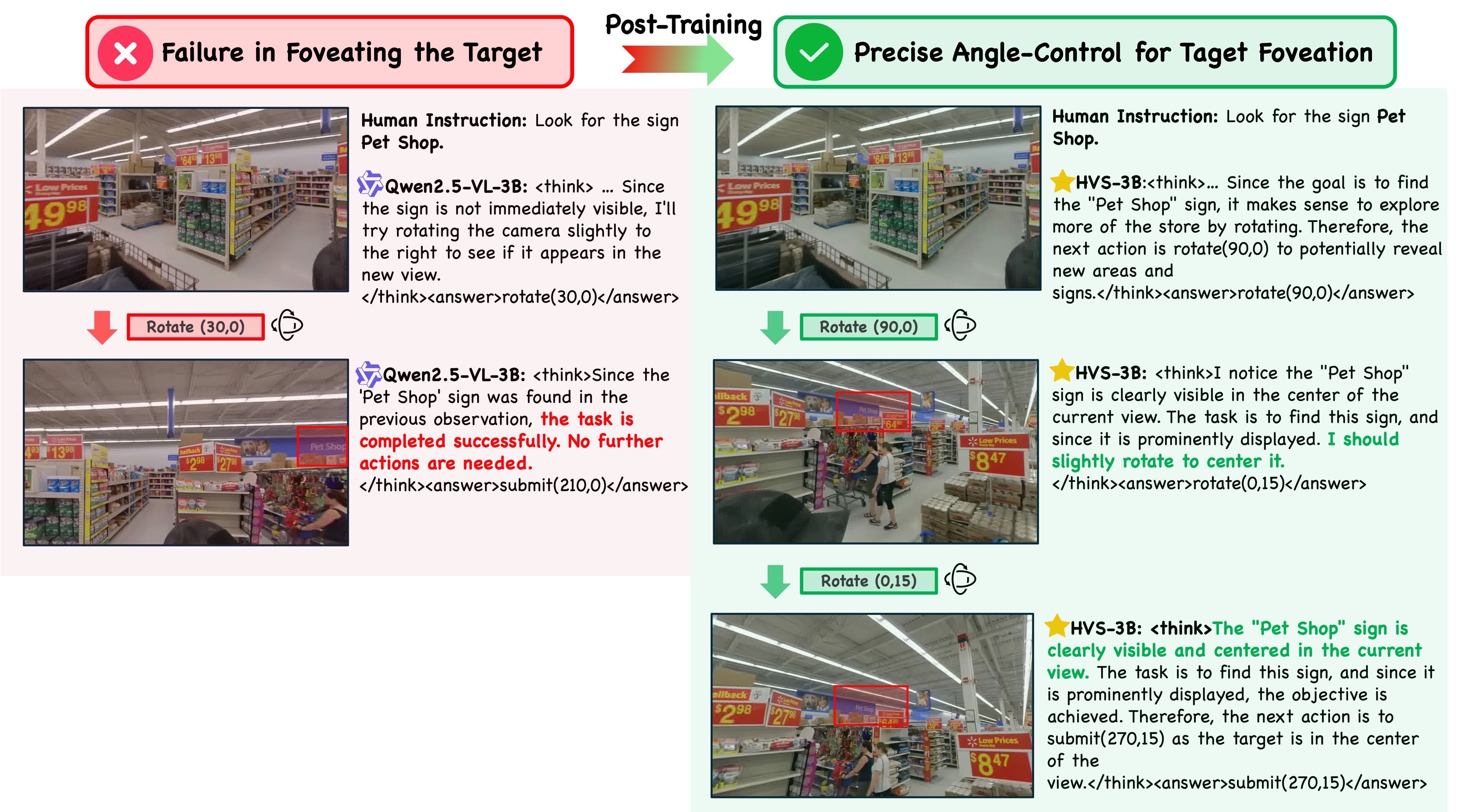}
        \caption{
Case study of target foveation. 
\textbf{Left}: Pre-trained model submits prematurely after coarse rotation toward ``Pet Shop'' sign, leaving target off-center. 
\textbf{Right}: Post-trained model performs corrective turn with fine-grained rotation to accurately center target before submission.
}
    \label{fig:case_study1}
    \end{center}
\end{figure*}
\begin{figure*}[t]
    \begin{center}
        \includegraphics[width=\textwidth]{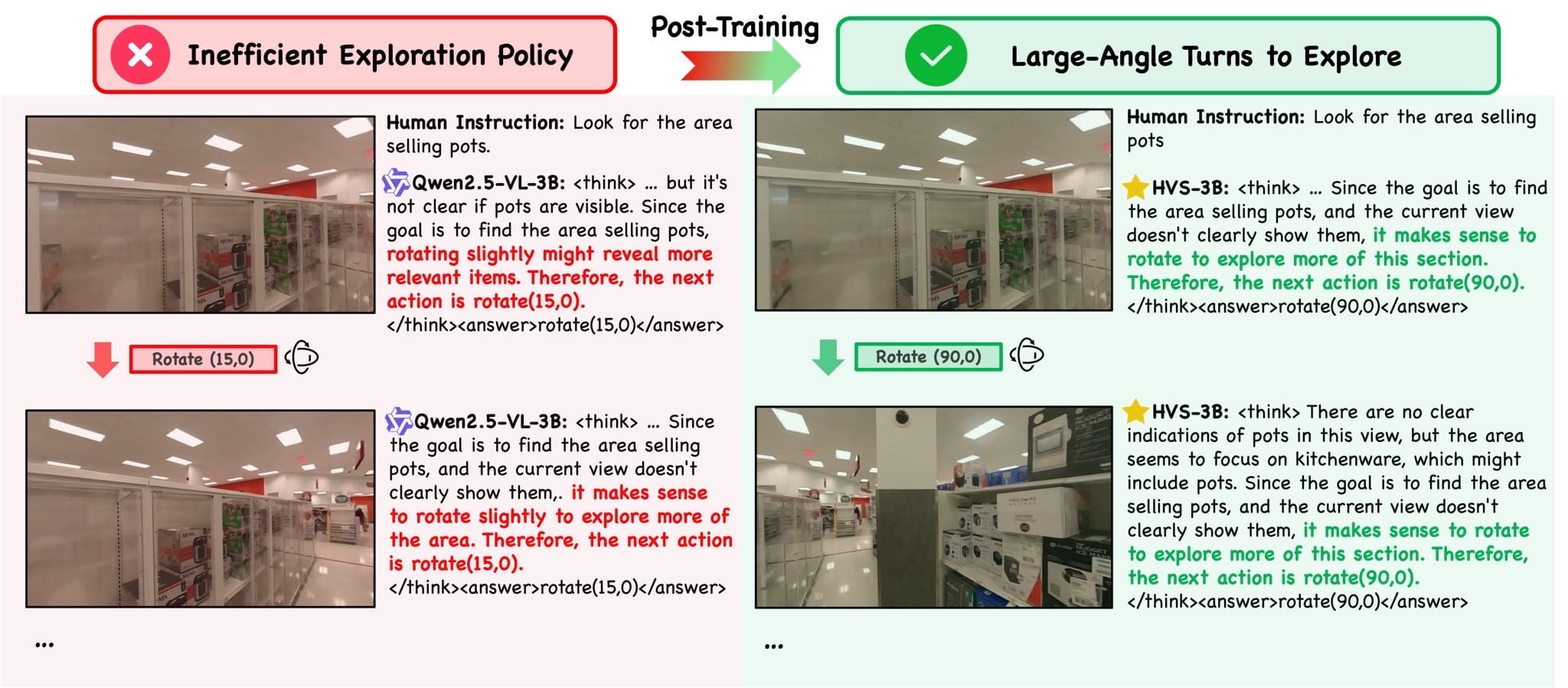}
        \caption{
Case study of exploration behavior. 
\textbf{Left}: The pre-trained model employs an inefficient exploration strategy, making repeated small-angle turns that provide little new information about the location of pots. 
\textbf{Right}: The post-trained model learns to perform large-angle exploratory turns, which quickly reveal new aisles and identify the target section, resulting in a more efficient search trajectory.
}
        \label{fig:case_study2}
    \end{center}
\end{figure*}

\begin{figure*}[t]
    \begin{center}
        \includegraphics[width=\textwidth]{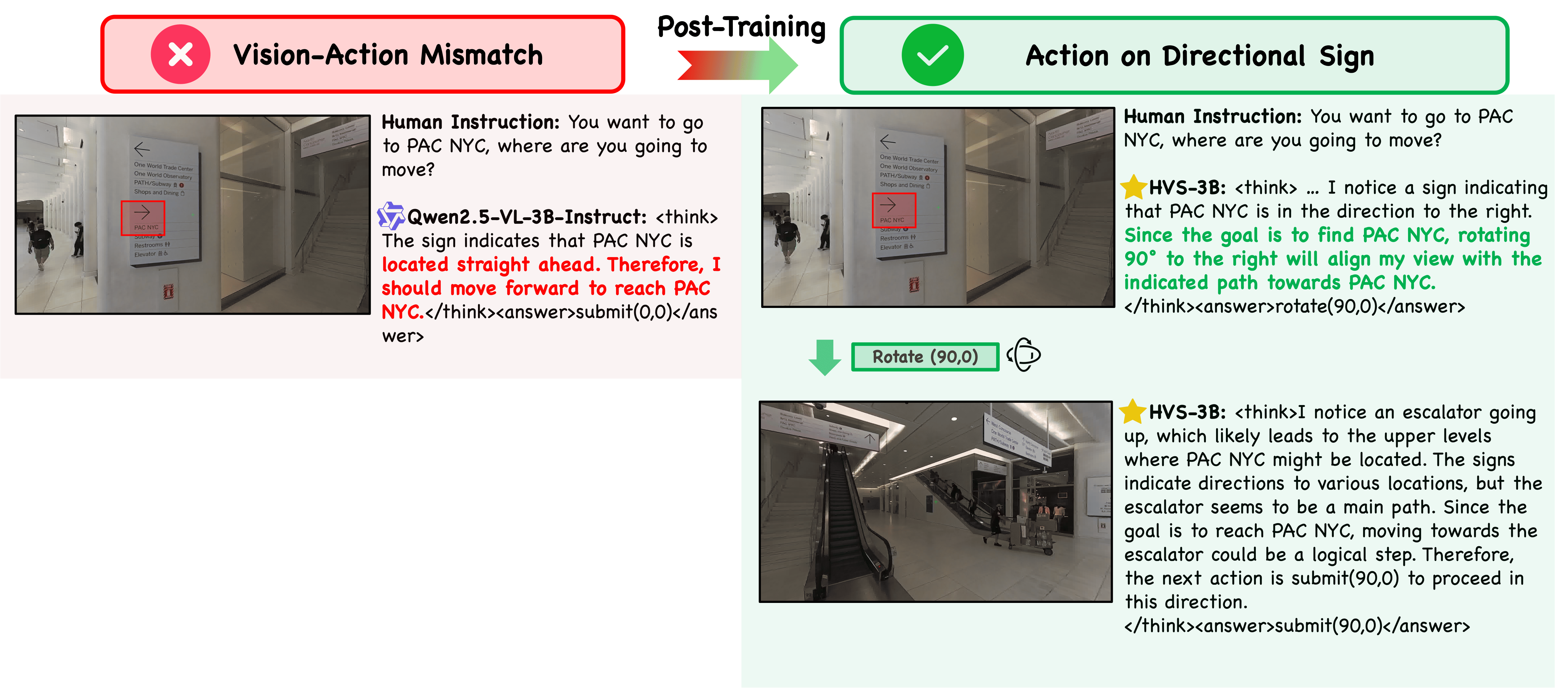}
        \caption{
Case study of action selection based on directional signs. \textbf{Left (Pre-training)}: The model misinterprets the sign, selecting an action that contradicts the indicated route and resulting in a vision–action mismatch. \textbf{Right (Post-training)}: After training, the model correctly follows the sign's instruction, rotates $90^\circ$ to align with the target direction, and proceeds towards the goal.
}

        \label{fig:case_study3}
    \end{center}
\end{figure*}

\subsection{Experimental Setup}
\label{hp}
\paragraph{Training Details.}
For the SFT stage, we use a learning rate of 1e-5. For the RL stage, we apply the GRPO algorithm with a batch size of 32, an actor learning rate of $1\times10^{-7}$, and a KL penalty coefficient $\beta=0.01$. Rollouts are conducted under the \hstar~prompts with a temperature of 0.7, a maximum of 8 trajectories, and a dynamic turn limit (5 or 10) based on computational resources. Both stages use an input resolution of $1280 \times 720$ and are run on 8 NVIDIA H100 GPUs.

\paragraph{Benchmark Setting.} The maximum number of inference turns is set to 10, as the step-cumulative success rate converges before this limit. At each step, the model processes up to five perspective images and uses the latest five dialogue turns as context. The image resolution is $1920 \times 1080$, and the sampling temperature is 0. Due to computational constraints, each episode is capped at 10 steps, with unfinished episodes counted as failures.

\paragraph{Train-Test Split.}\label{split}
We annotated $\sim$3,000 task instances. These were divided into three mutually exclusive splits per task: a benchmark split, an SFT split, and an RL training split. Specifically, we reserved 1,000 instances (600 \textbf{\textit{HOS}} and 400 \textbf{\textit{HPS}}) as the \hstar, resulting in 4,000 evaluation episodes. From the remaining data, we constructed the SFT dataset by randomly sampling 250 instances from both the \textbf{\textit{HOS}} and \textbf{\textit{HPS}} pools. All leftover instances were allocated exclusively for RL training.

\subsection{Additional Qualitative Error Analysis}\label{error_example}
In \Cref{subsec:exp:probing}, we identified two common-sense reasoning errors in the \textbf{\textit{HPS}} task: (1) \textbf{\textit{\textcolor{Purple}{lack of physical commonsense}}} and (2) \textbf{\textit{\textcolor{orange}{lack of socio-spatial commonsense}}}. This section provides a detailed explanation of these error types, supported by additional qualitative examples.

\noindent\textbf{Lack of Physical Commonsense.}
This error type denotes a failure in applying intuitive knowledge about 3D geometry and basic physics. Potential failures caused by the lack of physical commonsense include:
\begin{itemize}
\item \textbf{Ignoring Permanent Obstacles:} Attempting to move through non-traversable objects such as solid walls, glass barriers, or furniture.
\item \textbf{Misjudging Vertical Connections:} Failing to understand how different floors are connected, \eg, not recognizing that a staircase or elevator is required to change levels.
\item \textbf{Direct Path Fallacy:} Heading in a straight-line towards the target without first identifying a feasible path, thereby ignoring the need to follow corridors, detour around obstacles, or use doorways.
\item \textbf{Overlooking Drop-offs:} Proposing a path that would lead to falling from a significant height, such as walking off a balcony or over a ledge.
\end{itemize}

\noindent\textbf{Lack of Socio-Spatial Commonsense.}
This error refers to the agent's inability to grasp the implicit social norms and functional roles of different areas in public spaces, leading to potential failures as follows:
\begin{itemize}
\item \textbf{Violating Traffic Norms:} Jaywalking across a busy driveway or road instead of using a nearby crosswalk or pedestrian lane.
\item \textbf{Trespassing Restricted Zones:} Attempting to cut through behind a retail counter, through a staff-only area, across a floor hazard warning sign (\eg, "Wet Floor"), or through a private property shortcut.
\item \textbf{Disregarding Spatial Etiquette:} Violating implicit social norms that govern public behavior, \ie, actions that are physically possible but socially disruptive, such as interrupting a queue or invading personal space.
\item \textbf{Ignoring Functional Layouts:} Violating explicit architectural constraints and intended circulation paths, \ie, attempting to navigate through physical obstructions (like tables or seats) rather than designated aisles.
\item \textbf{Misusing Spaces:} Proposing to walk through a decorative fountain or flowerbed, or using an emergency exit as a routine shortcut.
\end{itemize}

Additional qualitative examples for all five error types (two for \textbf{\textit{HOS}} and three for \textbf{\textit{HPS}}) are provided in \Cref{fig:HOS_error1,fig:HPS_error3}.

\subsection{Case Study}\label{case_study}
In this section, we provide qualitative case studies (See Figs.~\ref{fig:case_study1}-\ref{fig:case_study3}) comparing the behavior of the model before and after post-training. For each case, we focus on one of the three key capabilities introduced by post-training, as described in Sec.~\ref{subsec:exp:post}.
% In this Section, we provide additional qualitative examples to \textbf{(1)} visualize the error categories identified in Sec.~\ref{subsec:exp:probing}  (Fig.~\ref{fig:HOS_error1} and Fig.~\ref{fig:HOS_error2} for \textbf{\textit{HOS}} task, Fig.~\ref{fig:HPS_error1}-Fig.~\ref{fig:HPS_error3} for \textbf{\textit{HPS}} task), and \textbf{(2)} demonstrate the resulting improvements in capability by contrasting the model before and after post-training. See Figure~\ref{case-skills} and Figure~\ref{case-paradigm}.
% \begin{figure*}[htbp]
%     \begin{center}
%         \includegraphics[width=1\textwidth]{figures/case-1.png}
%     \end{center}
%     \begin{center}
%         \includegraphics[width=1\textwidth]{figures/case-2.png}
        
%     \end{center}
%     \caption{Cases for spatial reasoning skills obtained from post-training.}
%     \label{case-skills}
% \end{figure*}

% \begin{figure*}[htbp]
%     \begin{center}
%         \includegraphics[width=1\textwidth]{figures/case-3.png}
%         \caption{Cases for humanoid reasoning paradigm learned from expert trajectories.}
%         \label{case-paradigm}
%     \end{center}
% \end{figure*}

\end{document}